\documentclass[11pt]{article}

\PassOptionsToPackage{table}{xcolor}
\usepackage[preprint]{acl}

\usepackage{times}
\usepackage{latexsym}
\usepackage[T1]{fontenc}
\usepackage[utf8]{inputenc}
\usepackage{microtype}
\usepackage{inconsolata}

\usepackage{amsmath}
\usepackage{amssymb}
\usepackage{amsfonts}

\usepackage{graphicx}
\graphicspath{{figures/}{./}}
\usepackage{booktabs}
\usepackage{multirow}
\usepackage{array}
\definecolor{lightblue}{RGB}{220,235,250}

\usepackage{xspace}
\usepackage{algorithm}
\usepackage{algpseudocode}
\usepackage{enumitem}
\usepackage{subcaption}
\usepackage[capitalize,noabbrev]{cleveref}
\usepackage{fontawesome5}

\newcommand{\method}{Agent-G$^2$\xspace}

\definecolor{myblue}{RGB}{30, 144, 255}
\definecolor{githubicon}{HTML}{24292F}
\definecolor{projecticon}{HTML}{00A98F}

\title{\raisebox{-0.22\height}{\includegraphics[height=1.25em]{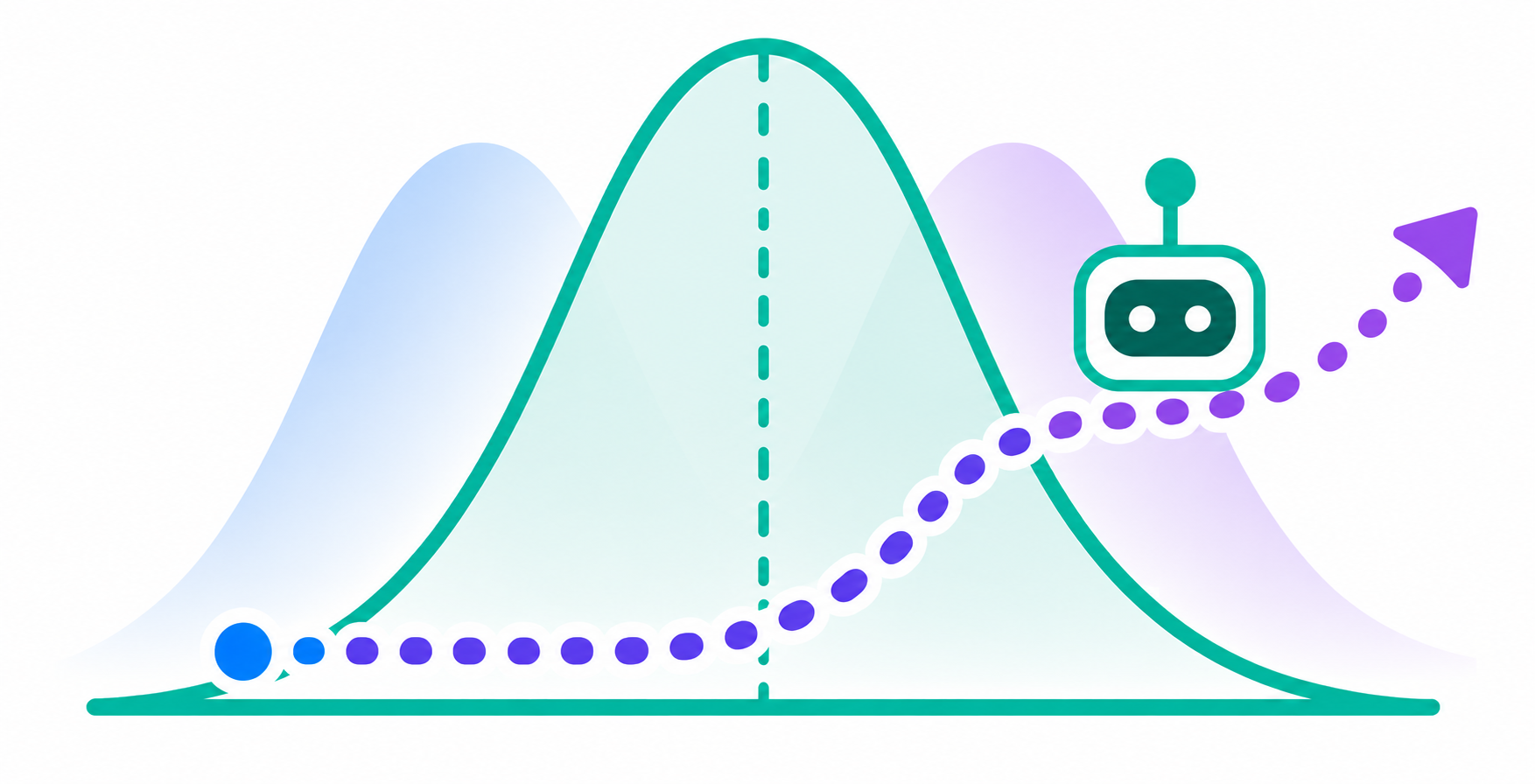}}\hspace{0.3em}%
Agent-G$^2$: \underline{G}aussian \underline{G}uidance for Agentic Reinforcement Learning}
\author{
  Zixuan Wang\textsuperscript{1,2}\thanks{\,Equal contribution.} \quad
  Yanrui Miao\textsuperscript{1,3}\footnotemark[1] \quad
  Zhengxi Lu\textsuperscript{1} \quad
  Teng Pan\textsuperscript{1,2} \\
  \bfseries
  Yiwen Qiu\textsuperscript{1} \quad
  Hongxing Li\textsuperscript{1} \quad
  Peng Qiu\textsuperscript{2} \quad
  Ruiqing Zhang\textsuperscript{2} \quad
  Yongliang Shen\textsuperscript{1}\thanks{\,Corresponding author.} \\[2pt]
  \mdseries
  \textsuperscript{1}Zhejiang University \quad
  \textsuperscript{2}Baidu Inc. \quad
  \textsuperscript{3}Shandong University \\
  \texttt{\{wang.zixuan, syl\}@zju.edu.cn} \\[3pt]
  \href{https://github.com/ZJU-REAL/Agent-G2}{%
    \textcolor{githubicon}{\faGithub}\,\texttt{Code}}
  \quad
  \href{https://zju-real.github.io/Agent-G2/}{%
    \textcolor{projecticon}{\faGlobe}\,\texttt{Project Page}}
  \quad
  \href{https://huggingface.co/collections/xiamoent/agent-g2}{%
    \raisebox{-0.22\height}{\includegraphics[height=1.15em]{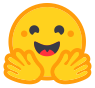}}\,\texttt{Models}}
}

\begin{document}
\maketitle

\begin{abstract}
Hint-based reinforcement learning addresses reward sparsity in long-horizon agentic tasks by retaining a prefix of an expert trajectory before each rollout, letting the policy explore from a state closer to success.
Its effectiveness hinges on the guidance depth: how much of the trajectory to keep.
Existing methods treat this depth as a deterministic scalar.
Scheduled approaches share one value across samples and ignore per-task heterogeneity; per-sample probing estimates it separately at the cost of extra rollouts.
We find that useful guidance occupies a band of depths whose informativeness profile is approximately Gaussian around the band center, rather than concentrating at a single optimal point.
We propose Agent-G$^2$, a Gaussian guidance framework that draws the depth per task from a Gaussian whose center and spread are estimated online from rollouts already collected for policy optimization, requiring no probe rollouts or learned depth predictor.
The center combines a global baseline with per-cluster difficulty, and the spread tracks within-cluster variance.
We evaluate Agent-G$^2$ on ALFWorld and WebShop on Qwen2.5-1.5B / 7B-Instruct.
Agent-G$^2$ outperforms the strongest hint-based, hint-free, and Aux-RL baselines on ALFWorld by 2.3 / 3.9 / 7.4 points at under one-third the rollout cost of per-sample probing.
\end{abstract}

\section{Introduction}
\label{sec:intro}

\begin{figure}[t] 
\centering 
\includegraphics[width=\linewidth]{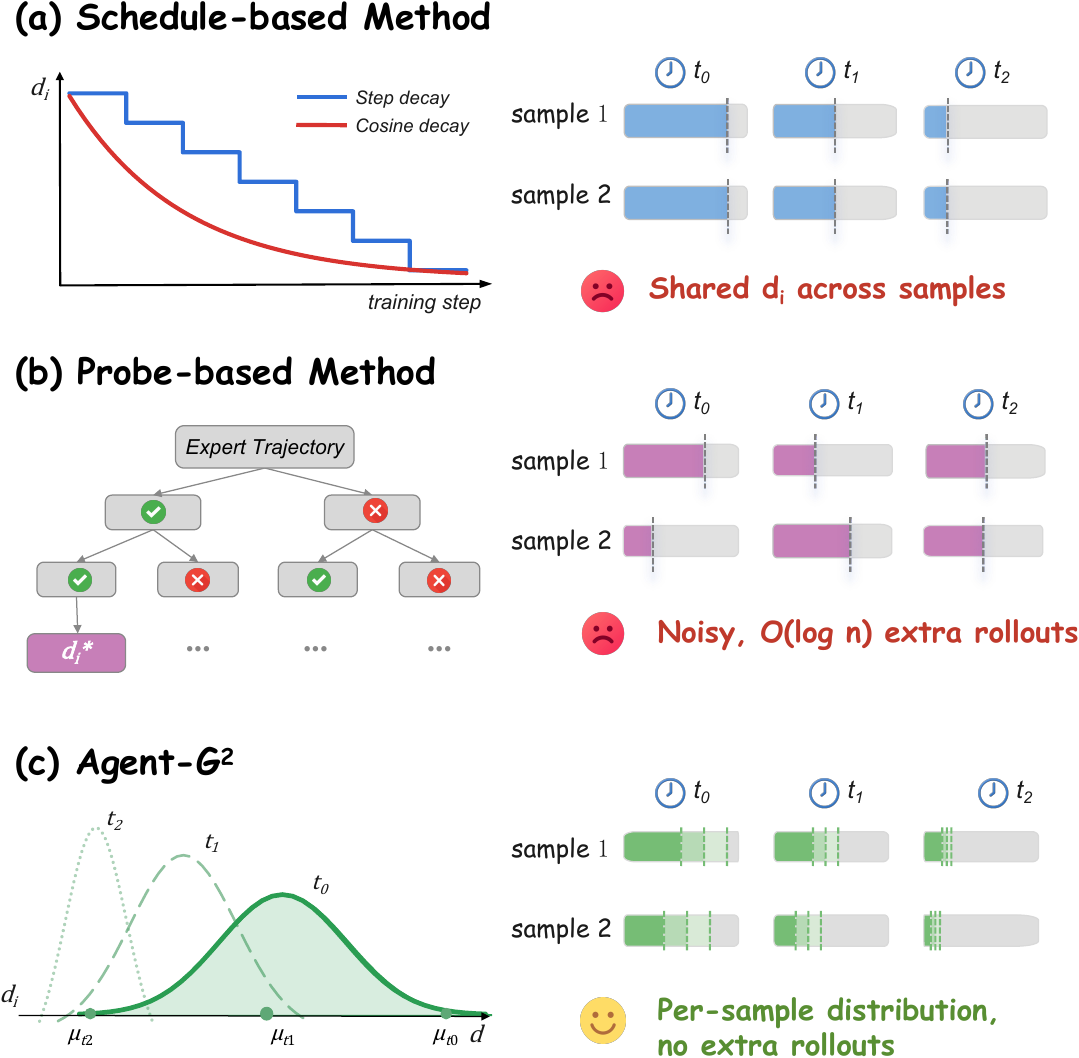} 
\caption{ 
\textbf{Hint-based RL paradigms} across training steps $t_0, t_1, t_2$. \textbf{(a)} Schedule-based: shared $d_t$ across samples. 
\textbf{(b)} Probe-based: $O(\log n)$ per-sample probes. 
\textbf{(c)} \method{} (ours): $d_i \sim \mathcal{N}(\mu_t, \sigma_t^2)$ per task, estimated online from existing rollouts.} 
\label{fig:preliminary} 
\end{figure}

LLM agents trained with reinforcement learning have made progress on long-horizon decision-making tasks~\citep{NEURIPS2023_77c33e6a,Zhang_2026}, including web navigation~\citep{Yao2022WebShopTS, zhou2023webarena, deng2023mind2web}, embodied control~\citep{ahn2022saycan, huang2022language, wang2025omniear}, and scientific experimentation~\citep{boiko2023autonomous, bran2023chemcrow}.
A recurring obstacle is \emph{reward sparsity}: tasks span dozens of sequential decisions, yet only a binary signal is issued at termination, and on-policy exploration from the initial state rarely reaches a successful terminal state.
Hint-based RL mitigates this obstacle by retaining a prefix of an expert trajectory before each rollout, so the policy explores from a state closer to success~\citep{xi2024r3, zhang2025stephint, su2025trapo}, directly addressing \emph{advantage collapse} in GRPO-style training~\citep{zhang2025bread, wang2025hint, li2025seele}.

The effectiveness of hint-based RL depends critically on the \emph{guidance depth}: how much of the expert trajectory to retain.
Too little guidance leaves few successful rollouts; too much saturates the reward and eliminates the contrast needed for advantage estimation.
Despite this sensitivity, existing methods uniformly treat depth as a \emph{deterministic scalar}, differing only in \emph{how} they select it (Figure~\ref{fig:preliminary}).
Schedule-based methods~\citep{xi2024r3, guo2025g2rpoa, huang2025prefixrft} derive one depth from the training step or batch-level feedback and share it across all samples, ignoring per-task heterogeneity.
Per-sample methods~\citep{su2025trapo, zhang2025bread, li2025seele, zhang2025stephint} estimate a separate depth through binary search or enumeration, but pay $O(\log n)$ extra rollouts per sample or $N\times$ the rollout budget.
Moreover, these methods have been developed almost exclusively on mathematical reasoning, where tasks share uniform structure; agentic tasks present a different challenge, as difficulty varies widely within a single batch (e.g., a two-step "Pick" vs.\ a twenty-step "Pick Two" in ALFWorld)~\citep{tu2026consensus, lian2026verbatim,li2026spatialevoselfevolvingspatialintelligence,chen2026knowubenchinteractiveproactivepersonalized}.

Our diagnostic in Section~\ref{sec:diagnostic} quantifies both failure modes: shared-depth schedulers place over 60\% of assignments outside the informative band (Figure~\ref{fig:diagnostic}(a)), while per-sample probing reduces this mismatch only by spending extra rollouts that remain too noisy under a GRPO-matched budget (Figure~\ref{fig:diagnostic}(b)).
Both families share a deeper assumption: that one optimal depth exists per task, and the goal is to pinpoint it.
Our analysis in Section~\ref{sec:gaussian-motivation} reveals that this assumption is mismatched to the problem structure: informative depths form a \emph{band} around the optimal point, with a training-signal profile that is unimodal, approximately symmetric, and well fit by a Gaussian ($\sigma{=}0.22$, $R^2{=}0.92$; Figure~\ref{fig:motivation}(b)).
\textbf{The right guidance depth is not a point to be found, but a neighborhood to be covered.}

This observation motivates \method{}, a Gaussian guidance framework for hint-based RL on long-horizon agentic tasks.
\method{} models guidance depth as a per-task Gaussian whose center and spread are estimated online through a global-local decomposition: a global baseline tracks the policy's overall progress, while per-cluster statistics adjust the center by task difficulty and widen the spread to match within-cluster variance.
For each task, one depth is drawn from its cluster's Gaussian and converted to a prefix length shared by all rollouts.
The same rollouts that update the policy also refresh the Gaussian parameters, achieving per-task depth variation at no extra rollout cost.
 
We evaluate \method{} on ALFWorld and WebShop with Qwen2.5-1.5B-Instruct and Qwen2.5-7B-Instruct. On ALFWorld, \method{} reaches $95.3\%$ overall success at 1.5B and $98.4\%$ at 7B, improving over the strongest hint-based RL baseline by $1.5$ and $2.3$ points and over the strongest hint-free RL baseline by $3.9$ points at both scales. On WebShop, \method{} obtains a $92.3$ reward score at both scales, with $78.9\%$ and $84.4\%$ final-purchase success. These gains hold across benchmarks, model scales, and horizon groups, without the probe-rollout overhead of per-sample search.
 
In summary, our contributions are as follows:
 
\begin{itemize}[nosep, leftmargin=*]
    \item We reveal the \emph{neighborhood structure} of guidance depth: informative depths form a band with approximately Gaussian training-signal profile, explaining the structural failure modes of both shared-depth schedulers and per-sample probing.
    \item We propose \method{}, a Gaussian guidance framework that estimates the per-task depth distribution online from rollout statistics via a global-local decomposition, requiring no probe rollouts or learned depth predictor.
    \item On ALFWorld and WebShop with Qwen2.5-1.5B / 7B-Instruct, we show that \method{} consistently outperforms the strongest non-probing hint-based, hint-free, and Aux-RL baselines on both benchmarks, at under one-third the rollout cost of per-sample probing.
\end{itemize}

\section{Preliminary Analysis}
\label{sec:diagnostic}
\begin{figure}[t]
    \centering
    \includegraphics[width=\linewidth]{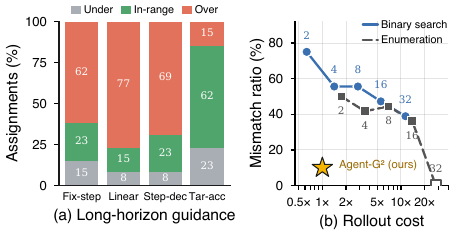}
    \caption{\textbf{Scalar-depth schedulers misallocate guidance.} (a) Shared-depth schedulers keep fewer than half of assignments in-range; (b) per-sample probing lowers the mismatch ratio $\rho$ only by spending extra rollouts.}
    \label{fig:diagnostic}
\end{figure}

We diagnose guidance-depth assignment on Qwen2.5-1.5B-Instruct / ALFWorld at step 50, where the policy's no-hint success rate is near 50\% and the spread of per-task informative depths is widest.
We sweep a depth grid $\mathcal{D}$ and estimate $p_i(d)$ from 32 rollouts per (task, depth) pair.
The in-range set of task $i$ is $\mathcal{R}_i = \{d \in \mathcal{D} : p_i(d) \in [0.4, 0.6]\}$ , where training is most informative~\citep{florensa2018automatic}; an assignment $d_i$ is under-guided if $p_i(d_i) < 0.4$ and over-guided if $p_i(d_i) > 0.6$.
The mismatch ratio $\rho$ is the fraction of tasks with $d_i \notin \mathcal{R}_i$.
Full details appear in Appendix~\ref{app:diagnostic}.

\paragraph{Mismatch in Shared-Depth Scheduling}
\label{sec:diagnostic-shared}

Shared-depth schedulers set one guidance depth from the training step or batch-level feedback and apply it to all tasks in a batch.
While avoiding additional probe rollouts, this scalar assignment ignores task-level heterogeneity.
Figure~\ref{fig:diagnostic}(a) shows that among the shared-depth schedulers we test, the three step-based schedulers place only 15\%--23\% of rollout groups inside the in-range band, with most assignments being over-guided.
Even the best shared scheduler, Target-acc, leaves 38\% of its assignments outside the in-range band.
This failure is structural rather than a tuning issue: a single depth cannot match tasks with different in-range bands.

\paragraph{Cost-precision Trade-off of Per-sample Probing}
\label{sec:diagnostic-probing}

Per-sample probing samples rollouts at candidate depths and selects the depth whose estimated success rate $\hat{p}_i$ is closest to $0.5$, removing the shared-depth assumption at the cost of additional probe rollouts.
Figure~\ref{fig:diagnostic}(b) quantifies this cost-precision trade-off.
With $M_{\text{probe}}{=}2$ rollouts per candidate depth, Binary Search still leaves 75\% of assignments mismatched.
Enumeration is more accurate but costlier: $M_{\text{probe}}{=}4$ reaches 47\% mismatch at $2\times$ the GRPO rollout budget, while near-zero mismatch requires $M_{\text{probe}}{=}32$ at $20\times$ the budget.
Thus, methods that try to pinpoint a single depth from noisy rollout estimates must trade additional rollout cost for selection error.

\paragraph{Modeling Depth as a Gaussian.}
The two failure modes above stem from a shared premise: that one optimal depth exists per task.
We now show that useful guidance has a richer structure.

\label{sec:gaussian-motivation}

\begin{figure}[t]
    \centering
    \includegraphics[width=\linewidth]{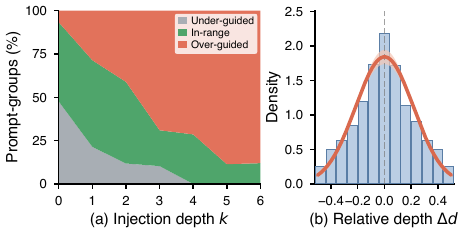}
    \caption{\textbf{Useful guidance forms a band around $d_i^\star$.} (a) Rollout success rate $p_i(d)$ vs.\ depth, with under-/in-range/over-guided regimes shaded; the in-range band spans several depths. (b) Bernoulli-variance informativeness aligned by $\Delta d = d - d_i^\star$, with a Gaussian fit ($\sigma{=}0.22$, $R^2{=}0.92$).}
    \label{fig:motivation}
\end{figure}

\noindent\textit{Observation 1: Useful guidance occupies a band.}
Figure~\ref{fig:motivation}(a) plots $p_i(d)$ against $d$, with the under-guided, in-range, and over-guided regimes shaded.
The in-range regime spans multiple neighboring depths rather than a single value, indicating that several nearby depths can keep rollouts in the just-learnable regime.
Thus, a scheduler that localizes one exact depth ignores a neighborhood of informative choices.

\noindent\textit{Observation 2: Informative signal is Gaussian-like.}
We define $d_i^\star = \arg\min_{d \in \mathcal{D}} |p_i(d) - 0.5|$ and align task-depth pairs by $\Delta d = d - d_i^\star$.
As a proxy for training informativeness we use the Bernoulli variance $p_i(d)(1-p_i(d))$, which peaks at $p_i{=}0.5$ and drops to $0$ at deterministic rollouts.
The peak at $\Delta d = 0$ follows from the definition of $d_i^\star$, but the symmetry and decay along $\Delta d$ do not: they reflect how $p_i(d)$ varies with depth.
The aligned profile (\cref{fig:motivation}(b)) is unimodal and approximately symmetric, with a Gaussian fit yielding $\sigma{=}0.22$ and $R^2{=}0.92$.
The Gaussian provides the two parameters our sampler needs: a center $\mu$ for the band location and a spread $\sigma$ for its width.
Both map onto the mean and variance of batch rollouts, so the sampler in \cref{sec:method} runs online without extra probing.
We discuss the choice of parametric family in \cref{app:gaussian-choice}.

\section{Method: \method{}}
\label{sec:method}

\begin{figure*}[t]
    \centering
    \includegraphics[width=\textwidth]{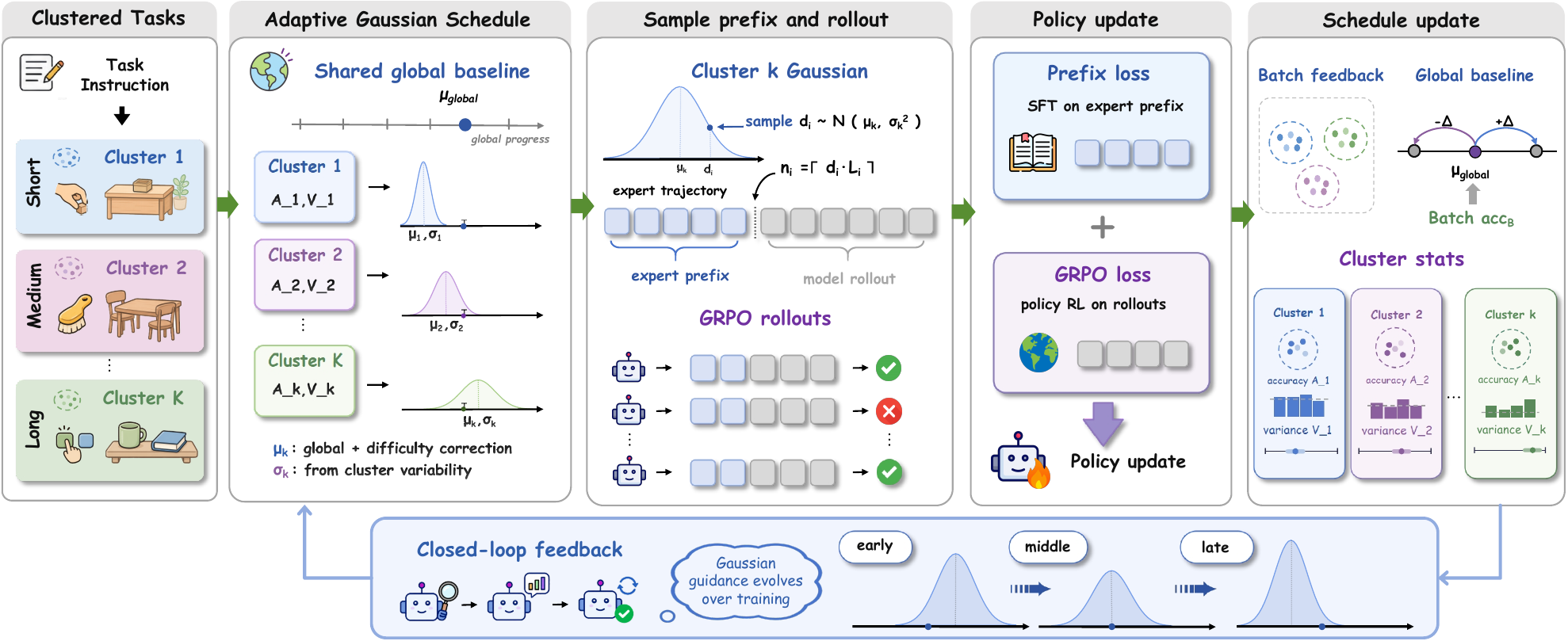}
    \caption{\textbf{\method{} pipeline.}
    \textbf{(1)} Tasks are clustered offline by difficulty.
    \textbf{(2)} A global baseline $\mu_{\text{global}}$ and per-cluster statistics $(A_k, V_k)$ form a per-task Gaussian $\mathcal{N}(\mu_i, \sigma_i^2)$.
    \textbf{(3)} One prefix length is drawn per task; $R$ rollouts run from the post-prefix state.
    \textbf{(4)} The policy is updated with prefix SFT plus GRPO loss.
    \textbf{(5)} The same rollouts refresh $\mu_{\text{global}}, A_k, V_k$ for the next batch.}
    \label{fig:method} 
\end{figure*}

Section~\ref{sec:diagnostic} showed that scalar-depth schedulers either ignore per-task heterogeneity or require costly per-sample probing, while informative depths form Gaussian-like profiles around task-specific effective depths.
\method{} translates these findings into a practical training framework (Figure~\ref{fig:method}): an adaptive Gaussian schedule derives per-task distributions from existing rollout statistics (\cref{sec:schedule}), sampling draws one depth per task (\cref{sec:sampling}), and the same rollouts update both the policy and the schedule (\cref{sec:training}).

\subsection{Problem Formulation.}
\label{sec:problem}
We consider LLM agents trained with reinforcement learning on long-horizon tasks with reward only at termination.
Each task $i$ has a natural-language instruction $q_i$ and one expert trajectory $\tau_i^\star = (a_{i,1}^\star, \dots, a_{i,L_i}^\star)$ of length $L_i$. 
A hint scheduler selects a guidance ratio $r_i \in [0,1]$, which is converted to a prefix length $n_i = \min(\lceil r_i L_i \rceil,\, L_i - 1)$. The first $n_i$ expert actions are executed in the environment to reach a post-prefix state, from which the policy $\pi_\theta$ runs $R$ independent rollouts. 
Each rollout $j \in \{1, \dots, R\}$ terminates with a binary reward $y_{i,j} \in \{0,1\}$, and $\hat{p}_i = \frac{1}{R}\sum_{j=1}^{R} y_{i,j}$ denotes the empirical success rate.
We optimize $\pi_\theta$ with GRPO \citep{shao2024deepseekmathpushinglimitsmathematical}, which computes a group-normalized advantage from the $R$ terminal rewards within each task.

\subsection{Adaptive Gaussian Schedule}
\label{sec:schedule}

\method{} assigns each task $i$ a Gaussian distribution $\mathcal{N}(\mu_i,\sigma_i^2)$ over the guidance ratio $r_i$.
Its center and spread are computed online from rollouts already collected for policy optimization.
The scheduler maintains three types of state: a global guidance baseline $\mu_{\text{global}}$ that tracks the overall guidance level, and for each offline cluster $\mathcal{C}_k$, EMA estimates $A_k$ and $V_k$ that summarize the level and dispersion of empirical success rates.
We use the standard midpoint target $p_{\mathrm{target}}=0.5$ for binary success feedback, following the curriculum principle that training is most informative near the success--failure boundary~\citep{florensa2018automatic,zhang2025stephint}.
All schedule states are initialized from weak priors that only determine the first-batch schedule, then continuously refreshed by rollout statistics.

\paragraph{Global baseline.}
The global baseline $\mu_{\text{global}} \in [0,1]$ tracks the guidance level required by the current policy.
For a training batch $\mathcal{B}$ with average success rate $\text{acc}_{\mathcal{B}} = \frac{1}{|\mathcal{B}|} \sum_{i \in \mathcal{B}} \hat{p}_i$, we shift $\mu_{\text{global}}$ by $\Delta$ in the direction that drives $\text{acc}_{\mathcal{B}}$ toward $p_{\mathrm{target}}$:
\begin{equation}
\small
\mu_{\text{global}} \leftarrow \mathrm{clip}\!\left(\mu_{\text{global}} + \mathrm{sign}(p_{\mathrm{target}} - \text{acc}_{\mathcal{B}})\,\Delta,\, 0,\, 1\right).
\label{eq:mu-global}
\end{equation}
Low batch success raises the baseline for deeper next-batch prefixes; high success lowers it.

\paragraph{Cluster statistics.}
The training set is partitioned offline into $K$ clusters $\{\mathcal{C}_k\}_{k=1}^K$ by each task's expert-trajectory length, which serves as a simple proxy for separating tasks of different difficulty levels, and $k(i)$ denotes the cluster of task $i$.
For each non-empty $\mathcal{B}_k = \mathcal{B} \cap \mathcal{C}_k$, the batch produces
\begin{equation}
\small
\begin{aligned}
\bar{p}_{\mathcal{B}_k} &= \frac{1}{|\mathcal{B}_k|} \sum_{i \in \mathcal{B}_k} \hat{p}_i,
\\
v_{\mathcal{B}_k} &= \frac{1}{|\mathcal{B}_k|} \sum_{i \in \mathcal{B}_k} \left(\hat{p}_i - \bar{p}_{\mathcal{B}_k}\right)^2,
\end{aligned}
\end{equation}
which are folded into $A_k$ and $V_k$ by EMA:
\begin{equation}
\small
\begin{aligned}
A_k &\leftarrow (1 - \alpha) A_k + \alpha \bar{p}_{\mathcal{B}_k},
\\
V_k &\leftarrow (1 - \alpha) V_k + \alpha v_{\mathcal{B}_k}.
\end{aligned}
\end{equation}
Clusters absent from the batch are left unchanged.
$A_k$ estimates the current success level of cluster $k$, while $V_k$ measures how unevenly tasks in the cluster respond to the current policy.

\paragraph{Per-task distribution.}
The center and spread for task $i$ in cluster $k{=}k(i)$ are
\begin{equation}
\small
\begin{aligned}
\mu_i &= \mathrm{clip}(\mu_{\mathrm{global}} + \lambda(p_{\mathrm{target}} - A_k),\, 0,\, 1), \\
\sigma_i &= \max(\gamma V_k,\, \sigma_{\min}).
\end{aligned}
\label{eq:per-task}
\end{equation}
A lower cluster success rate $A_k$ shifts $\mu_i$ toward deeper guidance; a higher within-cluster variance $V_k$ widens $\sigma_i$, giving the sampler broader coverage of the informative band identified in Section~\ref{sec:gaussian-motivation}.

\subsection{Per-Task Sampling and Rollout}
\label{sec:sampling}

Given $(\mu_i, \sigma_i)$ from the schedule, we draw one guidance ratio per task:
\begin{equation}
\small
z_i \sim \mathcal{N}(\mu_i, \sigma_i^2), \qquad
r_i = \mathrm{clip}(z_i,\, 0,\, 1).
\end{equation}
The sampled ratio determines the prefix length $n_i$, executed once to reach the post-prefix state shared by all $R$ rollouts of task $i$ (Section~\ref{sec:problem}).
Since $(\mu_i,\sigma_i)$ are induced by the statistics of cluster $k(i)$, tasks in the same cluster share Gaussian parameters, while independent sampling yields task-level depth variation without extra rollouts.
\subsection{Training}
\label{sec:training}

Each batch $\mathcal{B}$ closes the loop between policy optimization and schedule adaptation: rollouts generated from sampled prefixes drive both updates.

\paragraph{Policy update.}
We combine the GRPO loss $\mathcal{L}_{\mathrm{GRPO}}(\mathcal{B})$ with a teacher-forced loss on sampled expert prefixes, following common practice in hint-based RL~\citep{zhang2025bread, huang2025prefixrft}:
\begin{equation}
\mathcal{L}_{\mathrm{aux}}(\mathcal{B})
=
-\sum_{i \in \mathcal{B}} \sum_{t=1}^{n_i}
\log \pi_\theta(a_{i,t}^\star \mid s_{i,t}, q_i),
\end{equation}
where $s_{i,t}$ is the state reached after the first $t-1$ expert actions.
The full objective is
\begin{equation}
\mathcal{L}(\mathcal{B})
=
\mathcal{L}_{\mathrm{GRPO}}(\mathcal{B})
+
\eta\,\mathcal{L}_{\mathrm{aux}}(\mathcal{B}),
\label{eq:joint}
\end{equation}
with $\eta$ controlling the auxiliary weight.
The auxiliary term preserves imitation supervision on the sampled prefixes and is ablated in Section~\ref{sec:exp:ablation}.

\paragraph{Schedule update.}
The terminal rewards from the same rollouts refresh $\mu_{\text{global}}, A_k, V_k$ for the next batch as specified in Section~\ref{sec:schedule}.
Schedule adaptation therefore reuses rollouts already collected for policy optimization and introduces no probe rollouts or learned depth predictor.
Algorithm~\ref{alg:agentg2} summarizes the full per-batch loop.

\begin{algorithm}[t]
\small
\caption{Agent-G\textsuperscript{2} per-batch training loop.}
\label{alg:agentg2}
\begin{algorithmic}[1]
\Require Policy $\pi_\theta$, expert trajectories $\{\tau_i^\star\}$, clusters $\{\mathcal{C}_k\}_{k=1}^K$
\Require Hyperparameters $\Delta, \alpha, \lambda, \gamma, \sigma_{\min}, \eta, R$; $p_{\mathrm{target}}=0.5$
\State Initialize $\mu_{\mathrm{global}}\leftarrow 0.8$, $A_k\leftarrow 0$, $V_k\leftarrow 0$ \Comment{weak priors}
\For{each batch $\mathcal{B}$}
    \Statex \quad \textcolor{green!60!black}{$\triangleright$ \textit{Per-task schedule, sampling, and prefix execution}}
    \For{each task $i\in \mathcal{B}$ with $k=k(i)$}
        \State $\mu_i \leftarrow \mathrm{clip}(\mu_{\mathrm{global}}+\lambda(p_{\mathrm{target}}-A_k),0,1)$
        \State $\sigma_i \leftarrow \max(\gamma V_k, \sigma_{\min})$
        \State $z_i \sim \mathcal{N}(\mu_i,\sigma_i^2)$; $r_i \leftarrow \mathrm{clip}(z_i,0,1)$
        \State $n_i \leftarrow \min(\lceil r_i L_i \rceil, L_i - 1)$
        \State Execute $\tau_{i,1:n_i}^\star$ to reach state $s_i$
        \State Record $\tau_{i,1:n_i}^\star$ for $\mathcal{L}_{\mathrm{aux}}$
        \For{$j=1,\ldots,R$}
            \State Roll out $\pi_\theta$ from $s_i$; record $y_{i,j}$
        \EndFor
    \EndFor
    \Statex \quad \textcolor{green!60!black}{$\triangleright$ \textit{Policy update}}
    \State Update $\pi_\theta$: minimize $\mathcal{L}_{\mathrm{GRPO}}(\mathcal{B})+\eta\mathcal{L}_{\mathrm{aux}}(\mathcal{B})$
    \Statex \quad \textcolor{green!60!black}{$\triangleright$ \textit{Schedule feedback}}
    \State $\hat{p}_i \leftarrow \tfrac{1}{R}\sum_{j} y_{i,j}$ for $i\in\mathcal{B}$
    \State $\mathrm{acc}_{\mathcal{B}} \leftarrow \tfrac{1}{|\mathcal{B}|}\sum_{i\in\mathcal{B}} \hat{p}_i$
    \State Refresh $A_k, V_k, \mu_{\mathrm{global}}$ (Section~\ref{sec:schedule})
\EndFor
\State \Return $\pi_\theta$
\end{algorithmic}
\end{algorithm}

\section{Experiments}
\label{sec:exp}

\begin{table*}[t]
  \centering
  \footnotesize
  \setlength{\tabcolsep}{5pt}
  \providecommand{\best}[1]{\cellcolor[HTML]{FDECC4}\textbf{#1}}
  \providecommand{\second}[1]{\cellcolor[HTML]{D5E8D4}\underline{#1}}
  \caption{\textbf{Main results on ALFWorld and WebShop} (success rate, \%). \colorbox[HTML]{FDECC4}{\textbf{Best}} and \colorbox[HTML]{D5E8D4}{\underline{second-best}} are highlighted.}
  \begin{tabular}{@{}ll*{6}{>{\centering\arraybackslash}p{0.7cm}}c*{2}{>{\centering\arraybackslash}p{0.7cm}}@{}}
  \toprule
  \multirow{3}{*}{\textbf{Type}} & \multirow{3}{*}{\textbf{Method}}
    & \multicolumn{7}{c}{\textbf{ALFWorld}}
    & \multicolumn{2}{c}{\textbf{WebShop}} \\
  \cmidrule(lr){3-9}\cmidrule(lr){10-11}
  & & \multicolumn{2}{c}{\textbf{Short}}
    & \multicolumn{3}{c}{\textbf{Medium}}
    & \textbf{Long}
    & \multirow{2}{*}{\textbf{All}}
    & \multirow{2}{*}{\textbf{Score}}
    & \multirow{2}{*}{\textbf{Succ}} \\
  \cmidrule(lr){3-4}\cmidrule(lr){5-7}\cmidrule(lr){8-8}
  & & Pick & Look & Clean & Heat & Cool & Pick2 & & & \\
  \midrule

  \rowcolor{gray!10} \multicolumn{11}{l}{\textit{\textbf{Closed-Source Models}}} \\
  Prompting & GPT-4o~\citep{hurst2024gpt}                                    & 75.3 & 60.8 & 31.2 & 56.7 & 21.6 & 49.8 & 48.0  & 31.8 & 23.7 \\
  Prompting & Gemini-2.5-Pro~\citep{comanici2025gemini}                      & 92.8 & 63.3 & 62.1 & 69.0 & 26.6 & 58.7 & 60.3  & 42.5 & 35.9 \\
  \midrule

  \rowcolor{gray!10} \multicolumn{11}{l}{\textit{\textbf{Base: Qwen2.5-1.5B-Instruct}}} \\
  Prompting & Direct Prompt~\citep{yang2024qwen25}                                  & 5.9  & 5.5  & 3.3  & 9.7  & 4.2  & 0.0  & 4.1   & 23.1 & 5.2  \\
  Prompting & ReAct~\citep{yao2023reactsynergizingreasoningacting}            & 17.4 & 20.5 & 15.7 & 6.2  & 7.7  & 2.0  & 12.8  & 40.1 & 11.3 \\
  Prompting & Reflexion~\citep{shinn2023reflexionlanguageagentsverbal}       & 35.3 & 22.2 & 21.7 & 13.6 & 19.4 & 3.7  & 21.8  & 55.8 & 21.9  \\
  SFT       & Full SFT                                                       & 68.8 & 70.6 & 61.9 & 71.4 & 42.9 & 28.0 & 56.3  & 86.6 & 69.5   \\
  RL        & GRPO~\citep{shao2024deepseekmathpushinglimitsmathematical}     & 85.3 & 53.7 & 84.5 & 78.2 & 59.7 & 53.5 & 72.8  & 75.8 & 56.8 \\
  RL        & GiGPO~\citep{feng2025groupingrouppolicyoptimizationllm}        & 96.0 & 76.5 & 91.8 & 91.3 & 71.7 & 79.5 & 86.1  & 83.1 & 65.0 \\
  RL        & BEACON~\citep{wang2026beacon}                                  & \best{100}  & \second{88.2} & 86.7 & \best{100}  & 78.9 & \second{92.9} & 91.4  & 86.1 & 75.6 \\
  Aux-RL    & ETO~\citep{song2024eto}                                        & 73.6 & 46.3 & 66.2 & 68.3 & 62.8 & 55.6 & 66.4  & --   & --   \\
  Aux-RL    & RLVMR~\citep{zhang2025rlvmrreinforcementlearningverifiable}    & 95.2 & 78.8 & 91.2 & 90.2 & 83.9 & 77.6 & 87.9  & --   & --   \\
  Hint-RL   & Linear decay                                                   & 92.5 & 83.3 & 90.0 & 66.6 & 71.4 & 82.6 & 83.6  & 88.9   & 74.2   \\
  Hint-RL   & Cosine decay                                                   & \second{97.6} & 70.0 & 83.3 & 58.3 & 85.0 & 76.2 & 83.6  & 83.2   & 73.4   \\
  Hint-RL   & Step decay                                                     & 85.0 & \best{100}  & \second{95.0} & 83.3 & \second{95.2} & 86.9 & 89.8  & 89.1   & 74.2   \\
  Hint-RL   & Target acc                                                     & \best{100}  & 83.3 & \best{100}  & \best{100}  & 90.5 & 82.6 & \second{93.8}  & 86.4   & 75.0   \\
  Hint-RL   & Enumeration                                                    & \best{100}  & 81.8 & \best{100}  & 68.8 & 80.0 & 61.1 & 86.0  & \second{90.1}   & \second{78.1}   \\
  Hint-RL   & Binary Search~\citep{zhang2025bread}& 83.3 & 75.0 & 90.9 & 78.6 & \best{95.7} & 81.5 & 85.2  & 89.7   & 77.3   \\
  Hint-RL   & StepHint~\citep{zhang2025stephint}                             & 86.0   & 60.0   & 84.9   & 63.7   & 75.0   & 65.2   & 77.3    & 87.6   & 71.9   \\
  Hint-RL   & TRAPO~\citep{su2025trapo}                                      & 79.2   & 50.0   & 85.2   & 56.3   & 54.8   & 18.2   & 59.4    & 89.4   & 75.8  \\
  \rowcolor{lightblue!50} Hint-RL & \textbf{\method{} (Ours)}                & 96.8 & \best{100}  & \best{100}  & \second{92.9} & 84.2 & \best{94.7} & \best{95.3} & \best{92.3}   & \best{78.9}   \\
  \midrule

  \rowcolor{gray!10} \multicolumn{11}{l}{\textit{\textbf{Base: Qwen2.5-7B-Instruct}}} \\
  Prompting & ReAct~\citep{yao2023reactsynergizingreasoningacting}            & 48.5 & 35.4 & 34.3 & 13.2 & 18.2 & 17.6 & 31.2  & 46.2 & 19.5 \\
  SFT       & Full SFT                                                       & 75.0 & \best{100}  & 80.0 & 43.8 & 59.1 & 37.8 & 64.9  & 88.3   & 79.7   \\
  RL        & GRPO~\citep{shao2024deepseekmathpushinglimitsmathematical}     & 90.8 & 66.1 & 89.3 & 74.7 & 72.5 & 64.7 & 77.6  & 79.3 & 66.1 \\
  RL        & GiGPO~\citep{feng2025groupingrouppolicyoptimizationllm}        & 91.8 & 88.6 & 95.9 & 90.2 & 86.5 & 85.2 & 90.2  & 84.4 & 72.8 \\
  RL        & BEACON~\citep{wang2026beacon}                                  & \best{100}  & 81.8 & 96.3 & 92.9 & \second{94.7} & 90.0 & 94.5  & 87.7 & 79.7 \\
  Aux-RL    & ETO~\citep{song2024eto}                                        & 88.2 & 70.5 & 82.3 & 83.6 & 71.0 & 51.2 & 74.2  & --   & --   \\
  Aux-RL    & RLVMR~\citep{zhang2025rlvmrreinforcementlearningverifiable}    & 95.3 & 88.2 & 90.1 & 92.4 & 89.8 & 86.7 & 91.8  & --   & --   \\
  Hint-RL   & Linear decay                                                   & 97.5 & \second{92.9} & 66.7 & 85.2 & 87.5 & \second{92.0} & 85.2  & 91.1   & 82.8   \\
  Hint-RL   & Cosine decay                                                   & 97.1 & \best{100}  & 78.9 & 85.7 & 83.3 & 90.0 & 89.4  & 90.1   & 82.0   \\
  Hint-RL   & Step decay                                                     & 95.0 & \second{92.9} & 90.5 & 91.7 & \best{100}  & 80.0 & 91.4  & 92.1   & 83.6   \\
  Hint-RL   & Target acc                                                     & 97.5 & \best{100}  & 95.2 & 83.3 & 81.3 & \second{92.0} & 92.9 & 89.3   & 79.7   \\
  Hint-RL   & Enumeration                                                    & \best{100}   & \best{100}   & \second{96.4}   & \best{100}   & 90.5   & 86.7   & \second{96.1}    & \best{96.0}   & \best{89.8}   \\
  Hint-RL   & Binary Search~\citep{zhang2025bread}                           & \second{97.6}   & \best{100}   & 96.3   & 73.7   & 84.6   & \best{93.8}   & 92.2    & 91.2   & 83.6   \\
  Hint-RL   & StepHint~\citep{zhang2025stephint}                             & 97.3   & \best{100}   & 89.4   & \second{94.1}   & 79.1   & 91.7   & 91.4    & 90.9   & 82.0   \\
  Hint-RL   & TRAPO~\citep{su2025trapo}                                      & 89.3   & 66.7   & 88.5   & 90.0   & 68.8   & 44.8   & 75.0    & 90.6   & 77.3 \\
  \rowcolor{lightblue!50} Hint-RL & \textbf{\method{} (Ours)}                & \best{100} & \best{100}  & \best{100} & \best{100} & \best{100} & 91.7 & \best{98.4}  & \second{92.3}   & \second{84.4}   \\
  \bottomrule
  \end{tabular}
  \label{tab:alfworld_taskwise}
\end{table*}

\subsection{Setup}
\label{sec:exp:setup}

\paragraph{Benchmarks.}
We evaluate \method{} on two long-horizon agent benchmarks. 
ALFWorld~\citep{ALFWorld20} is a text-based embodied environment where the agent completes household tasks through multi-step interaction with sparse terminal rewards.
We follow the standard six task types and group them by horizon into \textit{Short} (Pick, Look), \textit{Medium} (Clean, Heat, Cool), and \textit{Long} (Pick2).
WebShop~\citep{Yao2022WebShopTS} is a web navigation environment where the agent searches, filters, and purchases products matching a natural-language specification, with reward issued at final purchase.

We compare \method{} against five families.
\textbf{(1) Prompting}: ReAct~\citep{yao2023reactsynergizingreasoningacting} and Reflexion~\citep{shinn2023reflexionlanguageagentsverbal} on the same base, with GPT-4o~\citep{hurst2024gpt} and Gemini-2.5-Pro~\citep{comanici2025gemini} as closed-source references.
\textbf{(2) Imitation}: Full SFT, supervised fine-tuning on entire expert trajectories.
\textbf{(3) RL without hints}: GRPO~\citep{shao2024deepseekmathpushinglimitsmathematical}, GiGPO~\citep{feng2025groupingrouppolicyoptimizationllm}, and BEACON~\citep{wang2026beacon}, trained with sparse terminal rewards and no expert prefix.
\textbf{(4) Aux-RL}: ETO~\citep{song2024eto} and RLVMR~\citep{zhang2025rlvmrreinforcementlearningverifiable}, which add auxiliary supervision beyond the terminal reward.
\textbf{(5) Hint-based RL}, subdivided into: \textit{schedule-based} methods (Linear, Cosine, Step decay, Target acc); \textit{search-based} methods (Binary Search, Enumeration); and \textit{end-to-end} methods (StepHint~\citep{zhang2025stephint}, TraPO~\citep{su2025trapo}).
All schedule- and search-based baselines share \method{}'s GRPO backbone, isolating prefix-depth allocation as the only variable.

\paragraph{Implementation.}
We instantiate \method{} on Qwen2.5-1.5B / 7B-Instruct~\citep{yang2024qwen25}.
\method{}-specific hyperparameters ($\Delta{=}0.1$, $\alpha{=}0.2$, $\gamma{=}1.0$) are fixed across both benchmarks without per-task tuning.
We report success rate (\%) on the held-out test set averaged over 5 seeds; full configurations appear in Appendix~\ref{app:impl_details}.

\subsection{Main Results}
\label{sec:exp:main}

\paragraph{Overall performance.}
On ALFWorld, \method{} reaches $95.3\%$ (1.5B) and $98.4\%$ (7B).
It outperforms RLVMR, the strongest method using auxiliary supervision, by $+7.4$ / $+6.6$, and Enumeration, the strongest probe-based baseline, by $+9.3$ / $+2.3$, without any auxiliary network or probe rollouts.
On WebShop, \method{} is the strongest non-probing method at both scales ($92.3$ reward score, $78.9\%$ / $84.4\%$ purchase success), and at 1.5B also outperforms Enumeration by $+2.2$ reward-score points.

\paragraph{Cross-scale robustness.}
\method{} remains effective as the backbone scales from 1.5B to 7B.
On ALFWorld, the margin over the strongest hint-based baseline grows from $+1.5$ to $+2.3$, so finer scheduling does not become redundant on stronger backbones.
On WebShop, purchase success climbs from $78.9\%$ to $84.4\%$, consistent with stronger action precision at scale.
The 1.5B model also surpasses all 7B non-probing baselines, indicating that schedule design substitutes for backbone scaling rather than only compensating for limited capacity.

\subsection{Analysis}
\label{sec:exp:analysis}

\begin{figure}[t]
    \centering
    \includegraphics[width=\linewidth]{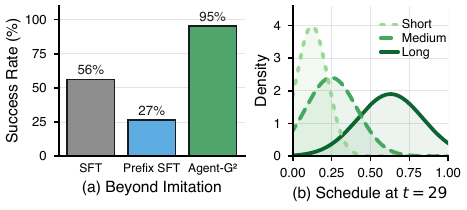}
    \caption{\textbf{Beyond imitation.} (a) ALFWorld success on Qwen2.5-1.5B: \method{} (95\%) far exceeds Full SFT (56\%) and Sampled-Prefix SFT (27\%). (b) Per-cluster (Short / Medium / Long) depth densities at $t{=}29$.}
    \label{fig:beyond_imitation}
\end{figure}

\paragraph{Beyond imitation.}
The gain of \method{} cannot be explained by imitation alone.
Full SFT reaches 56.3\%, and Sampled-Prefix SFT, which trains only on the prefix supervision pairs collected during \method{} training, reaches 26.6\%; both are far below \method{}'s 95.3\% (\Cref{fig:beyond_imitation}(a)).
The 68.4-point gap between Sampled-Prefix SFT and \method{} isolates the contribution of post-prefix RL: the missing ingredient is not copying expert tokens, but learning from sampled post-prefix states.

\begin{figure}[t]
    \centering
    \includegraphics[width=\linewidth]{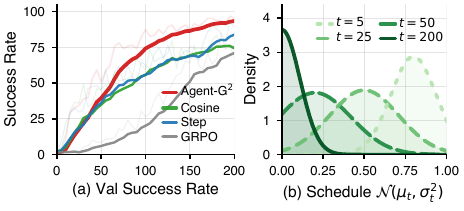}
    \caption{\textbf{Training dynamics of \method{}.} (a) Success rate vs.\ gradient steps for \method{} and the strongest scalar-depth baseline. (b) Evolution of the per-task guidance-depth distribution across training.}
    \label{fig:training_dynamics}
\end{figure}

\paragraph{Heterogeneous depths in the same batch.}
\method{} gains most when samples in the same batch require different prefix depths.
In Figure~\ref{fig:training_dynamics}(a), the convergence gap is largest between steps 50 and 150, the window in which the depth distribution in Figure~\ref{fig:training_dynamics}(b) is widest.
A scalar $d^{\mathrm{sched}}$ is brittle here: a single depth is too deep for easy samples and too shallow for hard ones, pushing most rollouts outside the informative band.
\method{} samples a neighborhood of depths instead, so both ends of the difficulty spectrum receive usable guidance.

\paragraph{Schedule emerges from the rollouts.}
The shape of the distribution in Figure~\ref{fig:training_dynamics}(b) is set by training itself.
At $t{=}5$ it favors deep prefixes, because the policy has yet to reach learnable states on its own.
Around $t{=}25$--$50$ it widens, as the cluster variance $V_k$ peaks while easy and hard clusters separate.
By $t{=}200$ it concentrates near zero, because the success rate has cleared the threshold of Equation~\eqref{eq:mu-global} on every cluster.
The mean follows global and cluster success, the spread follows $V_k$, and both come from rollouts already collected for policy optimization.
\Cref{fig:schedule_4panel} in Appendix~\ref{app:training_dynamics} samples this pattern across four training phases, with the Long cluster retaining a wider $\sigma_k$ for longer.

\begin{table}[t]
    \centering
    \footnotesize
    \setlength{\tabcolsep}{6pt}
    \caption{\textbf{Per-step training cost on Qwen2.5-1.5B / ALFWorld.} Median wall-clock seconds per gradient step. Ratios are relative to \method{}.}
    \begin{tabular}{@{}lccc@{}}
    \toprule
    \textbf{Method} & \textbf{Type} & \textbf{Cost/step (s)} & \textbf{Ratio} \\
    \midrule
    \rowcolor{lightblue!50} \textbf{\method{}} & Distribution & $88$ & $1.00\times$ \\
    \midrule
    Step decay    & Schedule & $57$ & $0.65\times$ \\
    Cosine decay  & Schedule & $60$ & $0.68\times$ \\
    Linear decay  & Schedule & $61$ & $0.69\times$ \\
    Target acc    & Schedule & $80$ & $0.91\times$ \\
    \midrule
    Enumeration   & Probe & $425$ & $4.83\times$ \\
    Binary Search & Probe & $285$ & $3.24\times$ \\
    \bottomrule
    \end{tabular}
    \label{tab:efficiency}
\end{table}

\paragraph{Training efficiency.}
We assess efficiency by convergence speed and per-step wall-clock cost.
Figure~\ref{fig:training_dynamics}(a) shows that \method{} reaches the final accuracy of scheduled baselines in roughly half the gradient steps.
Although \method{} incurs a modest per-step overhead over scheduled methods ($88$s vs. $57$--$80$s), it is substantially cheaper than probing-based methods ($285$--$425$s, or $3.24$--$4.83\times$ higher), which spend extra rollouts to estimate depths.
Therefore, by combining faster convergence with much lower cost than probing, \method{} achieves the best overall training efficiency.

\paragraph{Decomposing the learned schedule.}
\Cref{fig:beyond_imitation}(b) decomposes the learned schedule at $t{=}29$ by horizon group.
These patterns are not manually specified: each cluster only maintains rollout-derived statistics $A_k$ and $V_k$, and the separation emerges from training feedback.

\subsection{Ablation Studies}
\label{sec:exp:ablation}

\begin{table}[htbp]
  \centering
  \footnotesize
  \setlength{\tabcolsep}{4pt}
  \caption{\textbf{Ablation of \method{}} on ALFWorld / Qwen2.5-1.5B-Instruct. $\Delta$ is the change in All relative to the full method. Removing $\mathcal{L}_{\mathrm{GRPO}}$ is equivalent to Sampled-Prefix SFT.}
  \begin{tabular}{@{}lccr@{}}
  \toprule
  \textbf{Variant} & \textbf{Long} & \textbf{All} & $\boldsymbol{\Delta}$ \\
  \midrule
  \rowcolor{lightblue!50} \textbf{\method{}}                       & 94.7 & 95.3 & --                                                  \\
  \midrule
  \multicolumn{4}{l}{\textit{\textbf{Sampling form}}} \\
  \quad w/o sampling: $d_{i,j}{=}\mu_i, \sigma_i{=}0$                                   & 84.6   & 89.8   & \textcolor{red!75!black}{$\downarrow\,5.5$}    \\
  \quad Uniform sampling: $\mathcal{U}[\mu_i\!\pm\!\sigma_i]$             & 80.8   & 88.3   & \textcolor{red!75!black}{$\downarrow\,7.0$}    \\
  \midrule
  \multicolumn{4}{l}{\textit{\textbf{Adaptive moments}}} \\
  \quad w/o adaptive center: $\mu_i{=}\mu_{\mathrm{global}}$              & 79.2   & 91.4   & \textcolor{red!75!black}{$\downarrow\,3.9$}    \\
  \quad w/o adaptive spread: $\sigma_i{=}\sigma_{\min}$                   & 78.3   & 93.8   & \textcolor{red!75!black}{$\downarrow\,1.5$}    \\
  \quad w/o grouping: $K{=}1$                                            & 65.4   & 89.1   & \textcolor{red!75!black}{$\downarrow\,6.2$}    \\
  \midrule
  \multicolumn{4}{l}{\textit{\textbf{Loss components}}} \\
  \quad w/o $\mathcal{L}_{\mathrm{aux}}$                                   & 76.7 & 86.7 & \textcolor{red!75!black}{$\downarrow\,8.6$}         \\
  \quad w/o $\mathcal{L}_{\mathrm{GRPO}}$                                  & 10.0   & 26.6 & \textcolor{red!75!black}{$\downarrow\,68.7$}        \\
  \bottomrule
  \end{tabular}
  \label{tab:ablation}
\end{table}

\paragraph{Sampling drives the gain.}
Replacing the Gaussian draw $d_i\sim\mathcal{N}(\mu_i,\sigma_i^2)$ with the deterministic mean $d_i=\mu_i$ removes stochastic depth coverage and drops the overall success rate from $95.3\%$ to $89.8\%$.
This shows that a single-depth estimate cannot cover the informative band.
Replacing the Gaussian with a variance-matched uniform distribution reduces performance to $88.3\%$, confirming that the gain comes from coverage of the informative band, not the exact distributional shape.

\paragraph{Both center and spread matter.}
Removing the cluster-dependent center correction by setting $\mu_i=\mu_{\mathrm{global}}$ lowers the overall success rate to $91.4\%$, showing that a global baseline alone cannot capture difficulty variation across clusters.
Removing the adaptive spread by setting $\sigma_i=\sigma_{\min}$ reduces performance to $93.8\%$, with a pronounced drop on Long tasks ($94.7\%\rightarrow78.3\%$), indicating that dynamically adapting $\sigma_i$ is important for maintaining sufficient depth coverage, especially on hard long-horizon tasks.
Collapsing all tasks into a single cluster ($K=1$) drops the overall success rate to $89.1\%$, with the largest degradation on Long tasks, decreasing by $29.3$ points.
Together, these results show that both cluster-aware centering and adaptive spreading are necessary for handling heterogeneous task difficulty.

\paragraph{Auxiliary loss is supportive.}
Removing $\mathcal{L}_{\mathrm{aux}}$ from \cref{eq:joint} lowers the overall success rate by $8.6$ points, showing that prefix supervision helps stabilize training.
In contrast, removing $\mathcal{L}_{\mathrm{GRPO}}$ reduces the method to Sampled-Prefix SFT and drops performance to $26.6\%$, a $68.7$-point decrease.
This indicates that imitating sampled prefixes alone is insufficient; the main gain comes from GRPO learning on sampled post-prefix states, while $\mathcal{L}_{\mathrm{aux}}$ acts as a stabilizer for prefix-token imitation.

\section{Related Work}
\label{sec:related}

\paragraph{Hint-Guidance Strategies.}
Hint-based RL injects an expert trajectory prefix into rollouts to address reward sparsity, and methods differ primarily in how they choose the prefix depth $d$.
\emph{Schedule-based} methods set $d$ from the training step or batch success rate and share one value across samples~\citep{xi2024r3,huang2025prefixrft,guo2025g2rpoa,lu2026skill0}.
\emph{Per-sample} methods make $d$ instance-specific by probing each sample with extra rollouts~\citep{zhang2025stephint,zhang2025bread,wang2025hint,zhang2025scafgrpo,su2025trapo,li2025seele}.
In both lines, $d$ is a deterministic scalar that ignores per-task heterogeneity within a training context, and existing methods are evaluated almost exclusively on math reasoning.
Agent-G\textsuperscript{2} draws a per-task $d_i$ from a cluster-level Gaussian whose mean and variance are estimated from GRPO's rollout statistics, removing both the shared-scalar assumption and the probing overhead, and we evaluate it on long-horizon agentic benchmarks.

\paragraph{Auxiliary Supervision for Agentic RL.}
In agentic settings, prior work typically converts expert trajectories into supervision for an auxiliary model, rather than using them to seed rollouts directly.
These auxiliary signals take five forms: SFT targets for behavior cloning~\citep{zeng2024agenttuning,xi2025agentgym,qi2025webrl,bai2024digirl}, preference pairs for trajectory-level DPO~\citep{song2024eto,putta2024agentq,lai2024autowebglm,yuan2025agentr,lian2026uimopdmultiplatformonpolicydistillation}, value heads or Q-critics distilled from demonstrations~\citep{xiang2024retrospex,zhou2024archer,feng2024agile,gu2024webdreamer}, step-level process reward models~\citep{choudhury2025agentprm,wang2025sparl,xiong2024ipr,lu2026sdar}, and milestone-shaped rewards extracted from successful traces~\citep{wang2026beacon,zheng2026admire}.
Each route adds an auxiliary network and inherits its annotation, Monte-Carlo labeling, or reward-model fragility cost~\citep{zhang2025prmlessons,pan-etal-2026-coverrl}.
Agent-G\textsuperscript{2} uses the same expert trajectories directly as rollout-starting prefixes inside an on-policy GRPO loop, training no auxiliary model and requiring no further offline annotation.

\section{Conclusion}
\label{sec:conclusion}

We proposed Agent-G$^2$, a Gaussian guidance framework for hint-based reinforcement learning on long-horizon agentic tasks.
Rather than treating guidance depth as a deterministic scalar, Agent-G$^2$ draws it per task from a Gaussian whose center and spread are estimated online from rollouts collected for policy optimization, without learned depth predictor or extra probe rollouts.
Agent-G² is the strongest on ALFWorld and the strongest non-probing method on WebShop at both 1.5B and 7B scales, with 1.5B Agent-G² already surpassing 7B BEACON, showing that 
schedule design can substitute for backbone scaling.

\section*{Limitations}

Agent-G$^2$ relies on the availability of one expert trajectory per training task.
When such trajectories are unavailable or costly to obtain, the framework cannot be applied directly; extending it to weaker forms of supervision (e.g., suboptimal demonstrations or language hints) is an open direction.
 
The Gaussian parameterization is motivated by the empirical informativeness profile observed in Section~\ref{sec:gaussian-motivation}, which fits well on our two benchmarks.
Tasks with multimodal or heavily skewed depth profiles may benefit from richer distributional families, though our uniform-distribution ablation (Section~\ref{sec:exp:ablation}) suggests that the exact shape matters less than stochastic coverage of the informative band.
 
The difficulty clusters are defined offline by expert-trajectory length, which serves as a simple proxy for task horizon and difficulty.
Although effective in our experiments, this fixed partitioning does not adapt as the policy improves and task-relative difficulty changes.
Online difficulty estimation could further improve adaptivity.

\label{sec:limitations}

\section*{Ethics Statement}

We adhere to the ACL Code of Ethics and Code of Conduct. Our work uses only publicly available benchmarks and pretrained open-source models. Code and training scripts will be released under the MIT License upon publication. We used Claude for grammatical refinement; all scientific content and conclusions are the authors' own work.

\section*{Acknowledgements}
This work was supported by National Key Research and Development Project (No. 2024YFB3312900), National Natural Science Foundation of China (No. 62506332) and CCF-Baidu Open Fund.

\clearpage
\appendix
\section{Diagnostic Protocol Details}
\label{app:diagnostic}

\paragraph{In-range set and mismatch ratio.}
Let $\mathcal{D}$ denote the guidance-depth grid and $\mathcal{T}$ the diagnostic task set.
For each $(i, d) \in \mathcal{T} \times \mathcal{D}$, we estimate the per-depth success rate by enumeration with $M=32$ independent rollouts:
\begin{equation}
    \hat{p}_i(d) = \frac{1}{M} \sum_{m=1}^{M} \mathbf{1}\!\left[\text{rollout } m \text{ at depth } d \text{ succeeds}\right].
\end{equation}
Following \citet{florensa2018automatic}, the \emph{in-range set} of task $i$ is the subset of depths whose estimated success rate lies in the just-learnable band $[0.4, 0.6]$:
\begin{equation}
    \mathcal{R}_i = \left\{\, d \in \mathcal{D} : \hat{p}_i(d) \in [0.4, 0.6] \,\right\}.
\end{equation}
Depths with $\hat{p}_i(d) < 0.4$ are \emph{under-guided} and those with $\hat{p}_i(d) > 0.6$ are \emph{over-guided}.
Given a scheduler that assigns depth $d_i$ to each task $i$, the \emph{mismatch ratio} is the fraction of tasks placed outside their in-range set:
\begin{equation}
    \rho = \frac{1}{|\mathcal{T}|} \sum_{i \in \mathcal{T}} \mathbf{1}\!\left[d_i \notin \mathcal{R}_i\right].
\end{equation}
$\rho = 0$ corresponds to every task receiving an in-range depth, and $\rho = 1$ corresponds to every assignment being under- or over-guided.

\paragraph{Shared-depth schedulers.}
A shared scheduler outputs one depth $d_t$ per training step that is applied to every task in the batch, so the per-task assignment is $d_i = d_t$ for all $i$.
The stacked bars in \cref{fig:diagnostic}(a) report the empirical fraction of assignments in each of the three categories defined above: under-guided ($\hat{p}_i(d_t) < 0.4$), in-range ($d_t \in \mathcal{R}_i$), and over-guided ($\hat{p}_i(d_t) > 0.6$).

\paragraph{Mismatch across training checkpoints.}
The diagnostic in \cref{fig:diagnostic}(a) is computed at step $50$.
To verify that shared-depth mismatch is not specific to this checkpoint, we repeat the protocol at five additional Qwen2.5-1.5B / ALFWorld checkpoints (\cref{tab:mismatch_checkpoints}).
The Fix-step baseline, which holds $d_t$ constant within $20$-step windows, exceeds $48\%$ mismatch at every checkpoint, and the two step-decay schedules (Linear, Step-dec) stay near $50\%$.
Target-acc, the only shared scheduler with batch-level feedback, achieves the lowest mismatch ($39.8\%$ on average) but still misses the in-range band on roughly two in five tasks.
The structural mismatch of shared-depth scheduling thus persists across the training trajectory and is not a step-$50$ artifact.

\begin{table}[t]
\centering
\footnotesize
\setlength{\tabcolsep}{6pt}
\caption{\textbf{Mismatch ratio $\rho$ of shared-depth schedulers across five training checkpoints} on Qwen2.5-1.5B / ALFWorld. $n$ is the number of diagnostic tasks at that step.}
\label{tab:mismatch_checkpoints}
\begin{tabular}{@{}rrcccc@{}}
\toprule
\textbf{Step} & \textbf{$n$} & \textbf{Fix-step} & \textbf{Linear} & \textbf{Step-dec} & \textbf{Tar-acc} \\
\midrule
$10$  & $28$ & $67.9$ & $42.9$ & $42.9$ & $32.1$ \\
$20$  & $28$ & $64.3$ & $39.3$ & $39.3$ & $35.7$ \\
$40$  & $28$ & $75.0$ & $57.1$ & $57.1$ & $50.0$ \\
$80$  & $58$ & $48.3$ & $55.2$ & $55.2$ & $41.4$ \\
$160$ & $45$ & $71.1$ & $55.6$ & $55.6$ & $40.0$ \\
\midrule
\textbf{Avg} & --- & $\mathbf{65.3}$ & $\mathbf{50.0}$ & $\mathbf{50.0}$ & $\mathbf{39.8}$ \\
\bottomrule
\end{tabular}
\end{table}

\paragraph{Probing schedulers.}
A probing scheduler samples $M_{\mathrm{probe}}$ rollouts per visited depth on a candidate subset $\mathcal{D}' \subseteq \mathcal{D}$, forms the noisy estimate $\tilde{p}_i^{(M_{\mathrm{probe}})}(d)$, and outputs
\begin{equation}
    d_i = \arg\min_{d \in \mathcal{D}'} \bigl|\tilde{p}_i^{(M_{\mathrm{probe}})}(d) - 0.5\bigr|.
\end{equation}
$\mathcal{D}'$ is the full grid for Enumeration and the subset visited during the search for Binary Search.
The output $d_i$ is judged against the reference in-range set $\mathcal{R}_i$ via the mismatch ratio defined above.
\Cref{fig:diagnostic}(b) varies $M_{\mathrm{probe}} \in \{2, 4, 8, 16, 32\}$.
Scheduler implementations follow the hint-based RL baselines in \cref{tab:baselines}.

\paragraph{\method{}.}
For \method{} in \cref{fig:diagnostic}(b), we draw $d_i \sim \mathcal{N}(\mu_i, \sigma_i^2)$ using the schedule parameters reached at step $50$ ($5$ independent draws per task) and report the mismatch ratio averaged over draws and over tasks.
No probing rollouts are performed: $\mu_i$ and $\sigma_i$ are read from the rollouts already collected for the GRPO update at that step.

\begin{table}[t]
\centering
\footnotesize
\setlength{\tabcolsep}{5pt}
\caption{\textbf{\method{} hyperparameters.} Shared across ALFWorld and WebShop unless noted. Hint-based RL baselines reuse the \textit{Policy optimization}, \textit{Rollout}, and \textit{Training schedule} blocks.}
\label{tab:hyperparams}
\begin{tabular}{@{}lcr@{}}
\toprule
\textbf{Hyperparameter} & \textbf{Symbol} & \textbf{Value} \\
\midrule
\multicolumn{3}{l}{\textit{\method{}-specific}} \\
Global baseline step & $\Delta$ & $0.1$ \\
Cluster EMA rate & $\alpha$ & $0.2$ \\
Center scale & $\lambda$ & $1.0$ \\
Variance scale & $\gamma$ & $1.0$ \\
Variance floor & $\sigma_{\min}$ & $0.1$ \\
Aux.\ loss weight & $\eta$ & $0.5$ \\
Target success rate & $p_{\mathrm{target}}$ & $0.5$ \\
Initial baseline & $\mu_{\mathrm{global}}^{(0)}$ & $0.8$ \\
Difficulty clusters & $K$ & $3$ \\
\midrule
\multicolumn{3}{l}{\textit{Policy optimization}} \\
Optimizer & -- & AdamW \\
Peak learning rate & -- & $1{\times}10^{-5}$ \\
LR schedule & -- & cosine, $10\%$ warmup \\
AdamW $(\beta_1,\beta_2)$ & -- & $(0.9, 0.999)$ \\
Weight decay & -- & $0.01$ \\
GRPO clip ratio & $\epsilon$ & $0.2$ \\
KL penalty & $\beta_{\mathrm{KL}}$ & $0.01$ \\
Precision & -- & bfloat16 \\
\midrule
\multicolumn{3}{l}{\textit{Rollout}} \\
Tasks per step & $|\mathcal{B}|$ & $16$ \\
Rollouts per task & $R$ & $8$ \\
Max prompt length & -- & $7000$ \\
Max response length & -- & $512$ \\
Temperature (train/eval) & -- & $1.0$/$0.4$ \\
\midrule
\multicolumn{3}{l}{\textit{Training schedule}} \\
Steps (ALFWorld) & -- & $200$ \\
Steps (WebShop) & -- & $150$ \\
\bottomrule
\end{tabular}
\end{table}

\paragraph{Rollout-cost axis.}
The horizontal axis in \cref{fig:diagnostic}(b) reports $|\mathcal{D}'| \cdot M_{\mathrm{probe}} / R$, where $R = 8$ is the GRPO rollout budget per task.
\method{} is plotted at $1\times$ because its schedule is refreshed from these $R$ rollouts without any additional probing.

\paragraph{Reference depth $d_i^\star$.}
The per-task reference depth
\begin{equation}
    d_i^\star = \arg\min_{d \in \mathcal{D}} |\hat{p}_i(d) - 0.5|,
\end{equation}
with ties broken in favor of the smaller depth, is used only in \cref{fig:motivation}(b) to align profiles via $\Delta d = d - d_i^\star$.
It plays no role in the mismatch computation of \cref{sec:diagnostic}.

\paragraph{Gaussian fit.}
For \cref{fig:motivation}(b), we bin pairs by relative depth $\Delta d$ with width $0.1$ and fit $a \exp(-\Delta d^{2}/2\sigma^{2})$ to the binned mean Bernoulli variance by least squares.

\paragraph{Gaussian fit across training.}
\Cref{fig:motivation}(b) reports the fit at step $50$.
To check that the Gaussian shape is not specific to this checkpoint, we partition the training trajectory into three non-overlapping windows, pool aligned pairs $(\Delta d,\, p_i(d)(1-p_i(d)))$ within each window to enlarge the per-window sample size and reduce statistical noise in the fit, and refit the same form (\cref{tab:gaussian_stages}).
The fit quality stays high across all three windows ($R^2 \in [0.89, 0.94]$), and the fitted width $\sigma$ drifts only mildly from $0.198$ early to $0.231$ late, consistent with a gradually widening informative band as the policy improves and with the upward trend of the per-cluster $\sigma_k$ in \cref{fig:training_dynamics}(b).
The two-parameter Gaussian thus captures the informativeness profile throughout training, not only at the mid-training checkpoint used in the figure.

\begin{table}[t]
\centering
\footnotesize
\setlength{\tabcolsep}{6pt}
\caption{\textbf{Gaussian fit of the aligned informativeness profile across training windows} on Qwen2.5-1.5B / ALFWorld. Each window pools $(\Delta d,\, p_i(d)(1-p_i(d)))$ pairs collected within the listed checkpoint range; $n$ counts pairs entering the least-squares fit.}
\label{tab:gaussian_stages}
\begin{tabular}{@{}lccc@{}}
\toprule
\textbf{Training Stage} & $\sigma$ & $R^2$ & $n$ \\
\midrule
Early (ckpts $0$--$50$)    & $0.198$ & $0.906$ & $116$ \\
Mid   (ckpts $50$--$100$)  & $0.206$ & $0.888$ & $146$ \\
Late  (ckpts $100$--$150$) & $0.231$ & $0.936$ & $109$ \\
\bottomrule
\end{tabular}
\end{table}

\paragraph{Choice of Gaussian for the sampler.}
\label{app:gaussian-choice}
Our choice of Gaussian for the sampler in \cref{sec:method} rests on two design considerations rather than fit quality alone.
First, the Gaussian provides a simple two-moment parameterization whose mean and variance align directly with the two statistics that batch rollouts naturally produce: $\mu_i$ is driven by the per-cluster empirical success rate, and $\sigma_i$ by the per-cluster success-rate variance.
Other symmetric families require additional shape parameters (for instance, the degrees of freedom of Student-$t$ or the shape exponent of generalized normal) that cannot be read directly from rollout aggregates.
Second, Gaussian sampling on the bounded ratio $r_i \in [0, 1]$ admits a closed-form truncated form and is numerically stable; heavy-tailed alternatives such as Cauchy produce unbounded magnitudes that would require ad-hoc clipping.
The ablation in \cref{tab:ablation} confirms that on top of these design considerations, replacing the Gaussian with a variance-matched uniform distribution over the same band drops overall success by $7.0$ points and Long-horizon success by $13.9$ points.

\section{Implementation Details}
\label{app:impl_details}

\begin{figure*}[t]
    \centering
    \includegraphics[width=\linewidth]{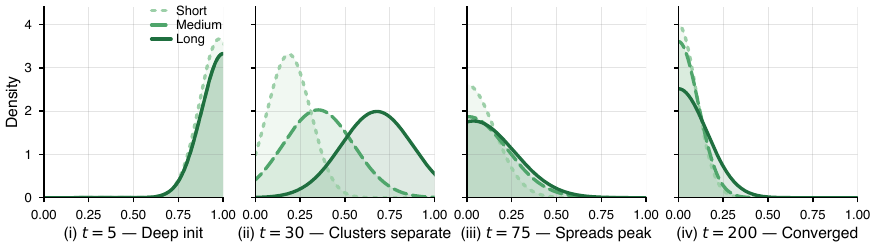}
    \caption{\textbf{Four-phase snapshot of the per-cluster Gaussian schedule} on Qwen2.5-1.5B / ALFWorld. Each panel shows the density $\mathcal{N}(\mu_k, \sigma_k^2)$ for the $K{=}3$ horizon clusters (Short, Medium, Long) at one training step. Their batch-weighted mixture yields the aggregate in \cref{fig:training_dynamics}(b).}
    \label{fig:schedule_4panel}
\end{figure*}

\begin{table}[t]
\centering
\footnotesize
\setlength{\tabcolsep}{5pt}
\caption{\textbf{Baseline configurations.} Hint-based RL baselines reuse the training blocks of \cref{tab:hyperparams}; only the depth-assignment rule differs.}
\label{tab:baselines}
\begin{tabular}{@{}ll@{}}
\toprule
\textbf{Method} & \textbf{Configuration} \\
\midrule
\multicolumn{2}{l}{\textit{Imitation}} \\
Full SFT & LR $1{\times}10^{-5}$, $|\mathcal{B}|{=}32$, $2$ epochs \\
\midrule
\multicolumn{2}{l}{\textit{RL without hints}} \\
GRPO & \cref{tab:hyperparams} with $d_i{=}0$ \\
GiGPO & authors' default \\
BEACON & authors' default \\
\midrule
\multicolumn{2}{l}{\textit{Hint-based RL: schedule}} \\
Linear & linear, $1.0\to0.0$ \\
Cosine & cosine, $1.0\to0.0$ \\
Step & step decay $0.1$ / $20$ steps \\
Target-acc & online to batch-acc $0.8$ \\
\midrule
\multicolumn{2}{l}{\textit{Hint-based RL: search}} \\
Binary Search & $O(\log|\mathcal{D}|)$ probes per task \\
Enumeration & exhaustive over $\mathcal{D}$ \\
\midrule
\multicolumn{2}{l}{\textit{Hint-based RL: end-to-end}} \\
StepHint & authors' default \\
TraPO & authors' default \\
\bottomrule
\end{tabular}
\end{table}

\paragraph{Hyperparameters.}
\Cref{tab:hyperparams} lists the full \method{} configuration.
On both benchmarks we build on the GiGPO codebase~\citep{feng2025groupingrouppolicyoptimizationllm}, adopting its training recipe for shared hyperparameters and inheriting its train/validation/test splits.

\paragraph{Baseline configurations.}
\Cref{tab:baselines} summarizes the trained baselines; prompting baselines run inference-time without further specification.
We use the authors' public implementations with default hyperparameters for GiGPO, BEACON, StepHint, and TraPO.

\paragraph{Expert trajectory source.}
The expert trajectories used by $\mathcal{L}_{\mathrm{aux}}$ are drawn from the same ALFWorld and WebShop expert pool as RLVMR~\citep{zhang2025rlvmrreinforcementlearningverifiable} and ETO~\citep{song2024eto}, so the underlying expert data is identical across the three methods.
The difference lies in usage: \method{} treats the prefix tokens only as a token-level imitation target inside the joint objective, without introducing any auxiliary network or additional supervision signal.
The gains over RLVMR and ETO in \cref{tab:alfworld_taskwise} therefore reflect algorithmic differences rather than any advantage in expert-data access.

\paragraph{Hardware and framework.}
We implement training in the verl framework~\citep{sheng2024hybridflow} and run all experiments on a single node with $8$ NVIDIA H800 GPUs; a Qwen2.5-1.5B-Instruct training run on ALFWorld ($200$ gradient steps) takes approximately $10$ hours.

\paragraph{Per-cluster training dynamics.}
\label{app:training_dynamics}

\begin{figure}[h]
    \centering
    \includegraphics[width=\linewidth]{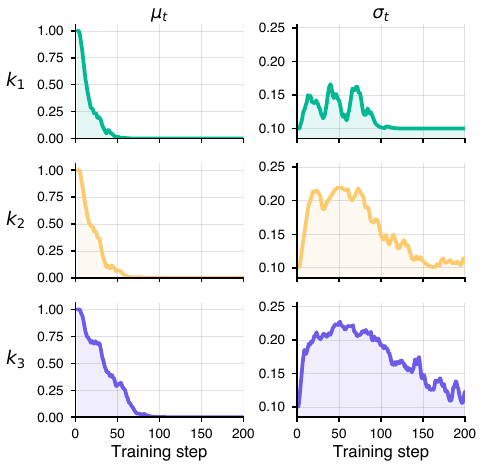}
    \caption{\textbf{Per-cluster schedule trajectories} on Qwen2.5-1.5B / ALFWorld. Each row shows one horizon cluster ($k_1$ Short, $k_2$ Medium, $k_3$ Long); columns plot $\mu_t$ and $\sigma_t$ over training steps.}
    \label{fig:cluster_schedule_dynamics}
\end{figure}

\Cref{fig:schedule_4panel} samples the per-cluster Gaussian schedule $\mathcal{N}(\mu_k, \sigma_k^2)$ at four training steps that mark distinct phases of training.
At $t{=}5$ (\emph{saturation}), $\mu_k$ stays near $1$ on every cluster ($0.98$, $1.00$, $1.00$ for Short, Medium, Long) with $\sigma_k\!\approx\!0.11$, because the policy cannot yet reach learnable states unaided.
At $t{=}30$ (\emph{separation}), the means split by horizon ($0.19$, $0.35$, $0.68$) and $\sigma_k$ widens with the cluster variance $V_k$.
At $t{=}75$ (\emph{peak spread}), the means contract toward $0$ on all clusters while $\sigma_k$ reaches its peak ($0.15$, $0.21$, $0.23$); the sampler draws a wide neighborhood around the shallow center to keep harder tasks in the just-learnable band.
By $t{=}200$ (\emph{collapse}), $\mu_k = 0$ on every cluster and $\sigma_k$ recedes toward $\sigma_{\min}{=}0.1$, with only Long retaining a wider spread ($0.16$), consistent with the slower convergence of long-horizon tasks.
\Cref{fig:cluster_schedule_dynamics} plots the full $(\mu_k, \sigma_k)$ trajectories across all $200$ training steps for the same three clusters.

\paragraph{Clustering design.}
\label{app:cluster_design}
\Cref{tab:cluster_design} sweeps the cluster count for length-quantile clustering and adds a random-assignment control on Qwen2.5-1.5B / ALFWorld.

\begin{table}[t]
\centering
\footnotesize
\setlength{\tabcolsep}{6pt}
\caption{\textbf{Clustering design ablation} on Qwen2.5-1.5B / ALFWorld (success rate \%). Length-quantile bins tasks by horizon; random samples a partition uniformly. $\pm$ values are $5$-seed standard deviations. $K{=}1$ and length-quantile $K{=}3$ are reproduced from \cref{tab:ablation}.}
\label{tab:cluster_design}
\begin{tabular}{@{}llc@{}}
\toprule
\textbf{Clustering signal} & $K$ & \textbf{All} \\
\midrule
None              & $1$  & $89.1$ \\
Length quantile   & $3$  & $95.3_{\pm 2.3}$ \\
Length quantile   & $5$  & $96.5_{\pm 1.9}$ \\
Length quantile   & $10$ & $86.2$ \\
Random            & $3$  & $89.4$ \\
\bottomrule
\end{tabular}
\end{table}

Length-quantile clustering is stable in the $K{=}3$--$5$ range: $K{=}3$ ($95.3_{\pm 2.3}$) and $K{=}5$ ($96.5_{\pm 1.9}$) overlap within run-to-run noise, so the difference between them is not significant under our seed budget.
Pushing to $K{=}10$ thins each cluster to under two tasks per batch under $|\mathcal{B}|{=}16$ and drops accuracy by $9.1$ points relative to $K{=}3$, as per-cluster sample size becomes inadequate for stable EMA estimation.
Random assignment at $K{=}3$ comes within $0.5$ points of $K{=}1$, so the gain from clustering relies on alignment with difficulty, not on partitioning itself.

We pick length as the alignment signal for two reasons.
(i) Length correlates with the difficulty of reaching the terminal reward without guidance, so it accelerates the convergence of $(A_k, V_k)$ relative to a difficulty-orthogonal signal.
(ii) Length needs no task-type annotation or semantic embedding, so the same rule applies to both benchmarks.

We fix $K{=}3$ across both ALFWorld and WebShop without per-benchmark tuning, because it matches the Short / Medium / Long taxonomy of \cref{sec:exp:setup} and keeps per-cluster sample size sufficient for stable EMA under $|\mathcal{B}|{=}16$.
The $6.2$-point $K{=}1$ vs.\ $K{=}3$ gap in \cref{tab:ablation} bounds what a richer signal could recover; in benchmarks where horizon and difficulty decouple, additional signals beyond length may help.


\begin{thebibliography}{56}
\providecommand{\natexlab}[1]{#1}

\bibitem[{Ahn et~al.(2022)Ahn, Brohan, Brown, Chebotar, Cortes, David, Finn, Fu, Gopalakrishnan, Hausman, Herzog, Ho, Hsu, Ibarz, Ichter, Irpan, Jang, Ruano, Jeffrey, Jesmonth, Joshi, Julian, Kalashnikov, Kuang, Lee, Levine, Lu, Luu, Parada, Pastor, Quiambao, Rao, Rettinghouse, Reyes, Sermanet, Sievers, Tan, Toshev, Vanhoucke, Xia, Xiao, Xu, Xu, Yan, and Zeng}]{ahn2022saycan}
Michael Ahn, Anthony Brohan, Noah Brown, Yevgen Chebotar, Omar Cortes, Byron David, Chelsea Finn, Chuyuan Fu, Keerthana Gopalakrishnan, Karol Hausman, Alex Herzog, Daniel Ho, Jasmine Hsu, Julian Ibarz, Brian Ichter, Alex Irpan, Eric Jang, Rosario~Jauregui Ruano, Kyle Jeffrey, and 26 others. 2022.
\newblock \href {https://arxiv.org/abs/2204.01691} {Do as i can, not as i say: Grounding language in robotic affordances}.
\newblock \emph{Preprint}, arXiv:2204.01691.

\bibitem[{Bai et~al.(2024)Bai, Zhou, Cemri, Pan, Suhr, Levine, and Kumar}]{bai2024digirl}
Hao Bai, Yifei Zhou, Mert Cemri, Jiayi Pan, Alane Suhr, Sergey Levine, and Aviral Kumar. 2024.
\newblock \href {https://arxiv.org/abs/2406.11896} {Digirl: Training in-the-wild device-control agents with autonomous reinforcement learning}.
\newblock \emph{Preprint}, arXiv:2406.11896.

\bibitem[{Boiko et~al.(2023)Boiko, MacKnight, and Gomes}]{boiko2023autonomous}
Daniil~A. Boiko, Robert MacKnight, and Gabe Gomes. 2023.
\newblock \href {https://arxiv.org/abs/2304.05332} {Emergent autonomous scientific research capabilities of large language models}.
\newblock \emph{Preprint}, arXiv:2304.05332.

\bibitem[{Bran et~al.(2023)Bran, Cox, Schilter, Baldassari, White, and Schwaller}]{bran2023chemcrow}
Andres~M Bran, Sam Cox, Oliver Schilter, Carlo Baldassari, Andrew~D White, and Philippe Schwaller. 2023.
\newblock \href {https://arxiv.org/abs/2304.05376} {Chemcrow: Augmenting large-language models with chemistry tools}.
\newblock \emph{Preprint}, arXiv:2304.05376.

\bibitem[{Chen et~al.(2026)Chen, Lu, Xu, Shao, Zhao, Tang, Du, Song, Liu, Yan, Zhang, Tan, Lu, Xiao, Zhuang, and Shen}]{chen2026knowubenchinteractiveproactivepersonalized}
Tongbo Chen, Zhengxi Lu, Zhan Xu, Guocheng Shao, Shaohan Zhao, Fei Tang, Yong Du, Kaitao Song, Yizhou Liu, Yuchen Yan, Wenqi Zhang, Xu~Tan, Weiming Lu, Jun Xiao, Yueting Zhuang, and Yongliang Shen. 2026.
\newblock \href {https://arxiv.org/abs/2604.08455} {Knowu-bench: Towards interactive, proactive, and personalized mobile agent evaluation}.
\newblock \emph{Preprint}, arXiv:2604.08455.

\bibitem[{Choudhury(2025)}]{choudhury2025agentprm}
Sanjiban Choudhury. 2025.
\newblock \href {https://arxiv.org/abs/2502.10325} {Process reward models for llm agents: Practical framework and directions}.
\newblock \emph{Preprint}, arXiv:2502.10325.

\bibitem[{Comanici et~al.(2025)Comanici, Bieber, Schaekermann, Pasupat, Sachdeva, Dhillon, Blistein, Ram, Zhang, Rosen, Marris, Petulla, Gaffney, Aharoni, Lintz, Pais, Jacobsson, Szpektor, Jiang, Haridasan, Omran, Saunshi, Bahri, Mishra, Chu, Boyd, Hekman, Parisi, Zhang, Kawintiranon, Bedrax-Weiss, Wang, Xu, Purkiss, Mendlovic, Deutel, Nguyen, Langley, Korn, Rossazza, Ramé, Waghmare, Miller, Byrd, Sheshan, Hadsell, Bhardwaj, Janus, Rissa, Horgan, Abdagic, Belenki, Allingham, Singh, Guidroz, Srinivasan, Schmit, Chiafullo, Elisseeff, Jha, Kolhar, Berrada, Ding, Si, Mallick, Och, Erell, Ni, Latkar, Yang, Sirkovic, Feng, Leland, Hornung, Wu, Blundell, Alvari, Huang, Yip, Deur, Liu, Surita, Duque, Damen, Jia, Guez, Mircea, Sinha, Magni, Stradomski, Marian, Galić, Chen, Husain, Singhal, Grewe, Aubet, Song, Blanco, Rechis, Ho, Munoz, Zheng, Hamrick, Mather, Taitelbaum, Rutherford, Lei, Chen, Shukla, Moreira, Doi, Isik, Shabat, Rogozińska, Kolipaka, Chang, Vušak, Venkatachary, Noghabi, Bharti, Jun, Zaks, Green,
  Challagundla, Wong, Mohammad, Hirsch, Cheng, Naim, Proleev, Vincent, Singh, Krikun, Krishnan, Ghahramani, Atias, Aggarwal, Kirov, Vytiniotis, Koh, Chronopoulou, Dogra, Ion, Tyen, Lee, Weissenberger, Strohman, Balakrishna, Rae, Velic, de~Liedekerke, Elyada, Yuan, Liu, Shani, Kishchenko, Alessio, Li, Song, Kwei, Jankowski, Pappu, Namiki, Ma, Tripuraneni, Cherry, Ikonomidis, Ling, Ji, Westberg, Wright, Yu, Parkinson, Ramaswamy, Connor, Yeganeh, Grover, Kenwright, Litchev, Apps, Tomala, Halim, Castro-Ros, Li, Boral, Sho, Yarom, Malmi, Klinghoffer, Lin, Ansell, S, Zhao, Zuo, Santoro, Cheng, Demmessie, Liu, Brichtova, Culp, Braun, Graur, Ng, Mehta, Phillips, Sundberg, Godbole, Liu, Katariya, Rim, Seyedhosseini, Ammirati, Valfridsson, Malihi, Knight, Toor, Lampe, Ittycheriah, Chiang, Yeung, Fréchette, Rao, Wang, Srivastava, Zhang, Rhodes, Brand, Weesner, Figotin, Gimeno, Fellinger, Marcenac, Leal, Marcus, Cotruta, Cabrera, Luo, Garrette, Axelrod, Baltateanu, Barker, Chen, Toma, Ingram, Riesa, Kulkarni, Zhang,
  Liu, Wang, Polacek, Wu, Hui, Reyes, Su, Barnes, Malhi, Siddiqui, Feng, Damaschin, Pighin, Steiner, Yang, Boppana, Ivanov, Kandoor, Shah, Mujika, Huang, Choquette-Choo, Patel, Yu, Creswell, Jerry, Liu, Barros, Razeghi, Roy, Culliton, Xiong, Pan, Strohmann, Powell, Seal, DeCarlo, Shyam, Katircioglu, Wang, Hardin, Odisho, Broder, Chang, Nair, Shtefan, O'Brien, Agarwal, Potluri, Goyal, Jhindal, Thakur, Stuken, Lyon, Toutanova, Feng, Wu, Horn, Wang, Cullum, Taubman, Shrivastava, Shi, Tomlinson, Patel, Tu, Oflazer, Pongetti, Yang, Taïga, Perot, Pierse, Han, Drori, Iturrate, Chakrabarti, Yeung, Dopson, ting Chen, Kulshreshtha, Guo, Pham, Schuster, Chen, Polozov, Xing, Zhou, Kacham, Kukliansky, Miech, Yaroshenko, Chi, Douglas, Fei, Blondel, Myla, Madmoni, Wu, Keysers, Kjems, Albuquerque, Yu, D'sa, Plantan, Ionescu, Elias, Gupta, Vuyyuru, Alcober, Zhou, Ji, Hartmann, Puttagunta, Song, Amid, Stefanoiu, Lee, Pucciarelli, Wang, Raul, Petrov, Tian, Anklin, Nti, Gomes, Schumacher, Vesom, Panagopoulos, Bousmalis, Andor,
  Jacob, Zhang, Rosgen, Kecman, Tung, Belias, Goodman, Covington, Wieder, Saxena, Davoodi, Huang, Maddineni, Roulet, Campbell-Ajala, Sessa, Xintian, Wu, Lai, Collins, Haig, Sakenas, Xu, Giustina, Shafey, Charoenpanit, Garg, Ainslie, Severson, Arenas, Pathak, Rajayogam, Feng, Bakker, Li, Wichers, Rogers, Geng, Li, Jagerman, Jia, Olmert, Sharon, Mauger, Mariserla, Ma, Mohabey, Kim, Andreev, Pollom, Love, Jain, Agrawal, Schroecker, Fortin, Warmuth, Liu, Leach, Blok, Girirajan, Aharoni, Uria, Sozanschi, Goldberg, Ionita, Ribeiro, Zlocha, Birodkar, Lachgar, Yuan, Choudhury, Ginsberg, Zheng, Dibb, Graves, Lokhande, Rasskin, Muraru, Quick, Tata, Sermanet, Chawla, Karo, Wang, Zhang, Keller, Dragan, Su, Chou, Liu, Tao, Prabhakara, Wilson, Liu, Wang, Evans, Du, Castaño, Prasad, Mahdy, Gerlach, Reid, Kahn, Zait, Pillai, Ulrich, Wang, Wassenberg, Farkash, Yalasangi, Wang, Bauza, Bucher, Liu, Yan, Leung, Sindhwani, Barnes, Singh, Jurin, Chang, Bhumihar, Eiger, Citovsky, Withbroe, Li, Xue, Santo, Stoyanov, Raimond, Zheng,
  Gao, Listík, Kwasiborski, Saputro, Ozturel, Mallya, Majmundar, West, Caron, Wei, Castrejon, Vikram, Ramachandran, Dhawan, Park, Smoot, van~den Driessche, Blau, Malik, Liang, Hirsch, dos Santos, Weinstein, van~den Oord, Lall, FitzGerald, Jiang, Yang, Webster, Elqursh, Pope, Rotival, Raposo, Zhu, Dean, Alabed, Tran, Gupta, Gleicher, Austin, Rosseel, Umekar, Das, Sun, Chen, Misiunas, Zhou, Di, Loo, Newlan, Li, Ramasesh, Xu, Chen, Gandhe, Soricut, Gupta, Hu, El-Sayed, Garcia, Brusilovsky, Chen, Bolt, Huang, Gurney, Zhang, Pritzel, Wilkiewicz, Seybold, Shamanna, Fischer, Dean, Gill, Mcilroy, Bhowmick, Selier, Yang, Cheng, Magay, Tan, Varma, Walder, Kocisky, Nakashima, Natsev, Kwong, Gog, Zhang, Dieleman, Jimma, Ryabtsev, Brahma, Steiner, Du, Žužul, Žanić, Raghavachari, Gierke, Zheng, Petrova, Dauphin, Liu, Kessler, Hand, Duvarney, Kim, Lee, Hussenot, Hui, Smith, Jain, Xia, Tomar, Amiri, Phan, Fuchs, Weyand, Tomasev, Cordell, Liu, Mallinson, Joshi, Crawford, Suggala, Chien, Fernando, Sanchez-Vargas,
  Williams, Crone, Luo, Karpov, Shan, Thurk, Strudel, Voigtlaender, Patil, Dozat, Khodaei, Singla, Ambroszczyk, Wu, Chang, Roark, Hegde, Ding, Filos, Wu, Pinto, Liu, Khanna, Pandey, Mcloughlin, Li, Haves, Zhou, Buchatskaya, Leal, de~Boursac, Akazawa, Anderson, Chen, Somandepalli, Liang, Goenka, Winkler, Grushetsky, Ding, Smith, Ye, Pont-Tuset, Li, Li, Golany, Wegner, Jiang, Barak, Shangguan, Vértes, Wong, Bornschein, Tudor, Bevilacqua, Schaul, Rawat, Zhao, Axiotis, Meng, McLean, Lai, Beattie, Kushman, Liu, Kutzman, Lang, Ye, Netrapalli, Mishra, Khan, Goel, Willoughby, Tian, Zhuang, Chen, Tsai, Kementsietsidis, Khare, Keeling, Xu, Waters, Altché, Popat, Mittal, Saxton, Badawy, Mathieu, Zheng, Zhou, Ranka, Shin, Duan, Salimans, Mihailescu, Shaham, Chang, Assael, Dikkala, Izzard, Cohen-Addad, Graves, Feinberg, Chung, Strouse, Karmon, Sharifzadeh, Ashwood, Pham, Blanton, Vasiloff, Barber, Geller, Zhou, Zubach, Huang, Zhang, Gupta, Young, Proskurnia, Votel, Gabeur, Barcik, Tripathi, Yu, Yan, Changpinyo,
  Pavetić, Coyle, Fujii, Mendez, Zhou, Rajamani, Hechtman, Cao, Juan, Tan, Dalibard, Du, Clay, Yao, Jia, Vijaykumar, Zhou, Bai, Hung, Pecht, Todorov, Khadke, Gupta, Lahoti, Autef, Duddu, Lee-Thorp, Bykovsky, Misiunas, Flennerhag, Thangaraj, McGiffin, Nado, Kunesch, Noever, Hertz, Liang, Stone, Palmer, Daruki, Pramanik, Põder, Kyker, Khan, Sluzhaev, Ritter, Ruderman, Zhou, Nagpal, Vodrahalli, Necula, Barham, Pavlick, Hartford, Shafran, Zhao, Mikuła, Eccles, Shimokawa, Garg, Vilnis, Chen, Shumailov, Lee, Abdelhamed, Xie, Cohen, Hlavnova, Malkin, Sitawarin, Lottes, Coquinot, Yu, Kumar, Zhang, Mahendru, Ahmed, Martens, Chen, Boag, Peng, Devin, Klimovskiy, Phuong, Vainstein, Xie, Ramabhadran, Howard, Yu, Goswami, Cui, Shleifer, Pinto, Yeh, Yang, Javanmardi, Ethier, Lee, Orbay, Kotecha, Bromberg, Shaw, Thornton, Rosenthal, Gu, Thomas, Gemp, Ayyar, Ushio, Selvan, Wee, Liu, Majzoubi, Yu, Abernethy, Liechty, Pan, Nguyen, Qiong, Hu, Perrin, Arora, Pitler, Wang, Shivakumar, Prost, Limonchik, Wang, Gao, Cour, Buch,
  Gui, Ivanova, Neubeck, Chan, Kim, Chen, Goyal, Chung, Liu, Su, Petrushkina, Shen, Joulin, Xu, Lin, Kulizhskaya, Chelba, Vasudevan, Collins, Bashlovkina, Lu, Fritz, Park, Zhou, Su, Tanburn, Sushkov, Rasquinha, Li, Prendki, Li, LV, Sharma, Fitoussi, Huang, Dai, Dao, Burrows, Prior, Qin, Pundak, Sjoesund, Khurshudov, Zhu, Webson, Kemp, Tan, Agrawal, Sargsyan, Cheng, Stephan, Kwiatkowski, Reid, Byravan, Michaely, Heess, Zhou, Goenka, Carpenter, Levskaya, Wang, Roberts, Leblond, Chikkerur, Ginzburg, Chang, Riachi, Chuqiao, Xu, Borsos, Pliskin, Pawar, Lustman, Kirkwood, Anand, Chaudhary, Kalb, Milan, Augenstein, Goldie, Prince, Raman, Sun, Xia, Cohen, Huo, Camp, Ellis, Zilka, Torres, Patel, Arora, Chan, Adler, Ayoub, Liang, Jamil, Jiang, Baumgartner, Sun, Karov, Akulov, Zheng, Cai, Fantacci, Rubin, Acha, Wang, D'Souza, Sathyanarayana, Dai, Rowe, Simanovsky, Goldman, Kuang, Pan, Rosenberg, Rojas-Esponda, Dutta, Zeng, Jurenka, Farquhar, Bansal, Iqbal, Roelofs, Joung, Beak, Ryu, Poplin, Wu, Alayrac, Buthpitiya,
  Ronneberger, Habtegebriel, Li, Cavallaro, Wei, Bensky, Denk, Ganapathy, Stanway, Joshi, Bertolini, Lo, Ma, Charles, Sampemane, Sahni, Chen, Askham, Gaddy, Young, Tan, Eyal, Bražinskas, Zhong, Wu, Epstein, Bailey, Hard, Lee, Goldshtein, Ruiz, Badawi, Lochbrunner, Kearns, Brown, Pardo, Weber, Yang, Jiang, Akin, Fu, Wainwright, Zou, Gaba, Manzagol, Kan, Song, Zainullina, Lin, Ko, Deshmukh, Jindal, Svensson, Tyam, Zhao, Kaeser-Chen, Baird, Moradi, Hall, Guo, Tsang, Liang, Pereira, Ganesh, Korotkov, Adamek, Thiagarajan, Tran, Chen, Tar, Jain, Dasgupta, Bilal, Reitter, Zhao, Vezzani, Gehman, Mehta, Beltrone, Dotiwalla, Guadarrama, Abbas, Karp, Georgiev, Ferng, Brockschmidt, Peng, Hirnschall, Verma, Bi, Xiao, Dabush, Xu, Wallis, Parker, Wang, Xu, Safarli, Tewari, Zhang, Kim, Gesmundo, Thomas, Levi, Chowdhury, Rao, Garst, Conway-Rahman, Ran, McKinney, Xiao, Yu, Agrawal, Stjerngren, Ionescu, Chen, Sharma, Chiu, Liu, Franko, Sanford, Cai, Michel, Ganapathy, Labanowski, Garrett, Vargas, Sun, Gale, Buschmann,
  Desjardins, Ghelani, Jain, Verma, Asawaroengchai, Eisenschlos, Harlalka, Kazawa, Metzler, Howland, Jian, Ades, Shah, Gangwani, Lee, Ring, Hernandez, Reich, Sinha, Sathe, Kovac, Gill, Kannan, D'olimpio, Sevenich, Whang, Kim, Sim, Chen, Zhang, Lall, Matias, Jia, Friesen, Nasso, Thapliyal, Perozzi, Yu, Shekhawat, Huda, Grabowski, Wang, Sreevatsa, Dib, Hassen, Schuh, Milutinovic, Welty, Quinn, Shah, Wang, Barth-Maron, Frye, Axelsson, Zhu, Ma, Giannoumis, Sedghi, Ye, Luan, Aydin, Chandra, Sampathkumar, Huang, Lavrenko, Eleryan, Hong, Hansen, Carthy, Samanta, Ćevid, Wang, Li, Voznesensky, Hoffman, Terzis, Sehwag, Fidel, He, Cai, He, Feng, Nikoltchev, Phatale, Chase, Lawton, Zhang, Ouyang, Tragut, Manshadi, Narayanan, Shen, Gao, Bolukbasi, Roy, Li, Golovin, Panait, Qin, Han, Anthony, Kudugunta, Patraucean, Ray, Chen, Yang, Bhatia, Talluri, Morris, Ražnatović, Brownfield, An, Peng, Kane, Zheng, Duduta, Kessinger, Noraky, Liu, Rong, Veličković, Rush, Goldin, Wei, Garlapati, Pantofaru, Kwon, Ni, Noland, Trapani,
  Beaufays, Roy, Chow, Turker, Cideron, Mei, Clark, Dou, Bošnjak, Leith, Du, Yazdanbakhsh, Nasr, Kwak, Sheth, Kaskasoli, Anand, Lakshminarayanan, Jerome, Bieber, Chu, Senges, Shen, Sridhar, Ndebele, Beyret, Mohamed, Chen, Freitag, Guo, Liu, Roit, Chen, Yan, Stone, Co-Reyes, Cole, Scellato, Azizi, Hashemi, Jin, Iyer, Valentine, György, Ahuja, Diaz, Lee, Clement, Kong, Garmon, Watts, Bhatia, Gupta, Miecnikowski, Vallet, Taly, Loper, Joshi, Atwood, Chick, Collier, Iliopoulos, Trostle, Gunel, Leal-Cavazos, Hrafnkelsson, Guzman, Ju, Forbes, Emond, Chauhan, Caine, Xiao, Zeng, Moufarek, Murphy, Meng, Gupta, Riedel, Das, Lawal, Narayan, Sosea, Swirhun, Friso, Neyshabur, Lu, Girgin, Wunder, Yvinec, Pyne, Carbune, Rijhwani, Guo, Doshi, Briukhov, Bain, Hitron, Wang, Gupta, Chen, Du, Zhang, Shah, Akula, Dylla, Kachra, Kuo, Zou, Wang, Xu, Zhu, Snyder, Menon, Firat, Mordatch, Yuan, Ponomareva, Blevins, Moore, Wang, Chen, Scholz, Dwornik, Lin, Li, Antognini, I, Song, Miller, Kalra, Raveret, Akerlund, Wu, Nystrom, Godbole,
  Liu, DeBalsi, Zhao, Liu, Caciularu, Lax, Khandelwal, Langston, Bailey, Lattanzi, Wang, Kovelamudi, Mondal, Guruganesh, Hua, Roval, Wesołowski, Ingale, Halcrow, Sohn, Angermueller, Raad, Stickgold, Lu, Kosik, Xie, Lillicrap, Huang, Zhang, Paulus, Farabet, Wertheim, Wang, Joshi, ling Ko, Wu, Agrawal, Lin, Sheng, Sung, Breland-King, Butterfield, Gawde, Singh, Zhang, Apte, Shetty, Hutter, Li, Salesky, Lebron, Kanerva, Paganini, Nguyen, Vallu, Peter, Velury, Kao, Hoover, Bortsova, Bishop, Jakobovits, Agostini, Agarwal, Liu, Kwong, Tavakkol, Bica, Greve, GP, Marcus, Hou, Duerig, Moroshko, Lacey, Davis, Amelot, Wang, Kim, Strinopoulos, Wan, Lan, Krishnan, Tang, Humphreys, Bai, Shtacher, Machado, Pang, Burke, Liu, Aravamudhan, Song, Hirst, Singh, Jou, Bai, Piccinno, Fu, Alazard, Meiri, Winter, Chen, Zhang, Heitkaemper, Lambert, Lee, Frömmgen, Rogulenko, Nair, Niemczyk, Bulyenov, Xu, Shemtov, Zadimoghaddam, Toropov, Wirth, Dai, Gollapudi, Zheng, Kurakin, Lee, Bullard, Serrano, Balazevic, Li, Schalkwyk, Murphy,
  Zhang, Sequeira, Datta, Agrawal, Sutton, Attaluri, Chiang, Farhan, Thornton, Lin, Choma, Nguyen, Dasgupta, Robinson, Comşa, Riley, Pillai, Mustafa, Golan, Zandieh, Lespiau, Porter, Ross, Rajayogam, Agarwal, Venugopalan, Shahriari, Yan, Xu, Tobin, Dubov, Shi, Recasens, Kovsharov, Borgeaud, Dery, Vasanth, Gribovskaya, Qiu, Mahdieh, Skut, Nielsen, Zheng, Yu, Bostock, Gupta, Archer, Rawles, Davies, Svyatkovskiy, Tsai, Halpern, Reisswig, Wydrowski, Chang, Puigcerver, Taege, Li, Schnider, Li, Dena, Xu, Telang, Shi, Zen, Kastner, Ko, Subramaniam, Kumar, Blois, Dai, Wieting, Lu, Zeldes, Xie, Hauth, Ţifrea, Li, El-Husseini, Abolafia, Zhou, Ding, Ghalebikesabi, Guía, Maksai, Ágoston Weisz, Arik, Sukhanov, Świetlik, Jia, Yu, Wang, Brand, Bloxwich, Kirmani, Chen, Go, Sprechmann, Kannen, Carin, Sandhu, Edkins, Nooteboom, Gupta, Maggiore, Azizi, Pritch, Yin, Gupta, Tarlow, Smith, Ivanov, Babaeizadeh, Goel, Kambala, Chu, Kastelic, Liu, Soltau, Stone, Agrawal, Kim, Soparkar, Tadepalli, Bunyan, Soh, Kannan, Kim, Chen,
  Halumi, Roy, Wang, Sercinoglu, Gibson, Bhatnagar, Sano, von Dincklage, Ren, Mitrevski, Olšák, She, Doersch, Jilei, Wang, Liu, Tan, Yakar, Warkentin, Ramirez, Lebsack, Dillon, Mathews, Cobley, Wu, Chen, Simon, Nath, Sainath, Bendebury, Julian, Mankalale, Ćurko, Zacchello, Brown, Sodhia, Howard, Caelles, Gupta, Evans, Bulanova, Katzen, Goldenberg, Tsitsulin, Stanton, Schillings, Kovalev, Fry, Shah, Lin, Upadhyay, Li, Radpour, Maggioni, Xiong, Haas, Brennan, Kamath, Savinov, Nagrani, Yacovone, Kappedal, Andriopoulos, Lao, Li, Rozhdestvenskiy, Hashimoto, Audibert, Austin, Rodriguez, Ruoss, Honke, Karkhanis, Xiong, Wei, Huang, Leng, Premachandran, Bileschi, Evangelopoulos, Mensink, Pavagadhi, Teplyashin, Chang, Xue, Tanzer, Goldman, Patel, Li, Wiesner, Zheng, Stewart-Binks, Han, Li, Luo, Lenc, Lučić, Xue, Mullins, Guseynov, Chang, Galatzer-Levy, Zhang, Bingham, Hu, Hartman, Ma, Griffith, Irpan, Radebaugh, Yue, Fan, Ungureanu, Sorokin, Teufel, Li, Anil, Paparas, Wang, Lin, Peng, Shum, Petrovic, Brady,
  Nguyen, Macherey, Li, Singh, Yenugula, Iinuma, Chen, Kopparapu, Stern, Dave, Thekkath, Perot, Kumar, Li, Xiao, Bilotti, Bateni, Noble, Lee, Vázquez-Reina, Salazar, Yang, Wang, Gruzewska, Rao, Raghuram, Xu, Ben-David, Mei, Dalmia, Zhang, Liu, Bansal, Pankov, Schwarcz, Burns, Chan, Sanghai, Liang, Liang, He, Stuart, Narayanan, Zhu, Frank, Fatemi, Sabne, Lang, Bhattacharya, Settle, Wang, McMahan, Tacchetti, Soares, Hadian, Cabi, Chung, Putikhin, Li, Chen, Tarango, Michalewski, Kazemi, Masoom, Sheftel, Shivanna, Vadali, Comanescu, Reid, Moore, Neelakantan, Sander, Herzig, Rosenberg, Dehghani, Choi, Fink, Hayes, Ge, Weng, Ho, Karro, Krishna, Thiet, Skerry-Ryan, Eppens, Andreetto, Sarma, Bonacina, Ayan, Nawhal, Shan, Dusenberry, Thakoor, Gubbi, Nguyen, Tsarfaty, Albanie, Mitrović, Gandhi, Chen, Epasto, Stephanov, Jin, Gehman, Amini, Weber, Behbahani, Xu, Allamanis, Chen, Ott, Sha, Jastrzebski, Qi, Greene, Wu, Toki, Vlasic, Shapiro, Kotikalapudi, Shen, Saeki, Xie, Cassirer, Bharadwaj, Kiyono, Bhojanapalli,
  Rosenfeld, Ritter, Mao, Oliveira, Egyed, Bandemer, Parisotto, Kinoshita, Pluto, Maniatis, Li, Guo, Ghiasi, Tarbouriech, Chatterjee, Jin, Katrina, Xu, Palomaki, Arnold, Sewak, Piccinini, Sharma, Albrecht, Purser-haskell, Vaswani, Chen, Wisniewski, Cao, Aslanides, Phu, Sieb, Agubuzu, Zheng, Sohn, Selvi, Andreassen, Subudhi, Eruvbetine, Woodman, Mery, Krause, Ren, Ma, Luo, Chen, Fan, Griffiths, Schuler, Li, Zhang, Sarr, Luo, Patana, Watson, Naboulsi, Collins, Sidhwani, Hoogeboom, Silver, Caveness, Zhao, Rodriguez, Deines, Bai, Griffin, Tagliasacchi, Xue, Babbula, Pang, Ding, Shen, Peake, Crocker, Raghvendra, Swisher, Han, Singh, Wu, Pchelin, Munkhdalai, Alon, Bacon, Robles, Bulian, Johnson, Powell, Ferreira, Li, Benzing, Velimirović, Soyer, Kong, Tony, Nguyên, Yang, Liu, van Amersfoort, Gillick, Sun, Rauschmayr, Zhang, Zhan, Zhou, Frolov, Yang, Vnukov, Rouillard, Li, Mandhane, Fallen, Venkataraman, Hu, Brennan, Lee, Chang, Sundermeyer, Pan, Ke, Tong, Fabrikant, Bono, Gu, Foley, Mao, Delakis, Bhaswar,
  Frostig, Li, Zipori, Hope, Kozlova, Mishra, Djolonga, Schiff, Merey, Briakou, Morgan, Wan, Hassidim, Skerry-Ryan, Sengupta, Jasarevic, Kallakuri, Kunkle, Brennan, Lieber, Mansoor, Walker, Zhang, Xie, Žužić, Chukwuka, Druinsky, Cho, Yao, Naeem, Butt, Kim, Jia, Jordan, Lelkes, Kurzeja, Wang, Zhao, Over, Chakladar, Prasetya, Jha, Ganapathy, Cong, Shroff, Saroufim, Miryoosefi, Hammad, Nasir, Xi, Gao, Maeng, Hora, Cheng, Haghani, Lewenberg, Lu, Matysiak, Raisinghani, Wang, Baugher, Sukthankar, Giang, Schultz, Fiedel, Chen, Lee, Dey, Zheng, Paul, Smith, Ly, Wang, Bansal, Perz, Ricco, Blank, Keshava, Sharma, Chow, Lad, Jalan, Osindero, Swanson, Scott, Ilić, Li, Jonnalagadda, Soudagar, Xiong, Batsaikhan, Jarrett, Kumar, Shah, Lawlor, Waters, Graham, May, Ramos, Lefdal, Cankara, Cano, O'Donoghue, Borovik, Liu, Grimstad, Alnahlawi, Tsihlas, Hudson, Grigorev, Jia, Huang, Igwe, Lebedev, Tang, Krivokon, Garcia, Tan, Jia, Stys, Vashishth, Liang, Venkatraman, Gu, Kementsietsidis, Zhu, Jung, Bai, Hosseini, Ahmed,
  Gupta, Yuan, Ashraf, Nigam, Vasudevan, Awasthi, Gilady, Mariet, Eskander, Li, Hu, Garrido, Schlattner, Zhang, Saxena, Dević, Muralidharan, Murthy, Zhou, Choi, Wongpanich, Wang, Shah, Xu, Huang, Spencer, Chen, Cohan, Wang, Tompson, Wu, Haroun, Li, Huergo, Yang, Yin, Wendt, Bendersky, Chaabouni, Snaider, Ferret, Jindal, Thompson, Xue, Bishop, Phal, Sharma, Sung, Radhakrishnan, Shomrat, Ingle, Vij, Gilmer, Istin, Sobell, Lu, Nottage, Sadigh, Willcock, Zhang, Xu, Brown, Lee, Wang, Zhu, Tay, Kim, Gutierrez, Sharma, Xian, Seo, Cui, Pochernina, Baetu, Jastrzębski, Ly, Elhawaty, Suh, Sezener, Wang, Yuen, Tucker, Cai, Yang, Wang, Muzio, Qian, Yoo, Lockhart, McKee, Guo, Mehrotra, Mendonça, Mehta, Ben, Tekur, Mu, Zhu, Krakovna, Lee, Maschinot, Cevey, Choe, Bai, Srinivasan, Gasaway, Young, Siegler, Holtmann-Rice, Piratla, Baumli, Yogev, Hofer, van Hasselt, Grant, Chervonyi, Silver, Hogue, Agarwal, Wang, Singh, Flynn, Lipschultz, David, Bellot, Yang, Le, Graziano, Olszewska, Hui, Maurya, Parotsidis, Chen, Oguntebi,
  Kelley, Baddepudi, Mauerer, Shaw, Siegman, Yang, Shetty, Roy, Song, Stokowiec, Burnell, Savant, Busa-Fekete, Miao, Ghosh, MacDermed, Lippe, Dektiarev, Behrman, Mentzer, Nguyen, Wei, Verma, Knutsen, Dasari, Yan, Mitrichev, Wang, Shejwalkar, Austin, Sunkara, Potti, Virin, Wright, Liu, Riva, Pot, Kochanski, Le, Balasubramaniam, Dhar, Liao, Bloniarz, Shukla, Cole, Lee, Zhang, Kafle, Vashishtha, Mahmoudieh, Chen, Hoffmann, Srinivasan, Lago, Shalom, Wang, Elabd, Sharma, Oh, Kothawade, Le, Monteiro, Yang, Alarakyia, Geirhos, Mincu, Garnes, Kobayashi, Mariooryad, Krasowiak, Zhixin, Lai, Mourad, Wang, Bu, Aharoni, Chen, Goyal, Zubov, Bapna, Dabir, Kothari, Lamerigts, Cao, Shar, Yew, Kulkarni, Mahaarachchi, Joshi, Zhu, Lichtarge, Zhou, Muckenhirn, Selo, Vinyals, Chen, Brohan, Mehta, Cogan, Wang, Geri, Ko, Chen, Viola, Shivam, Wang, Elish, Popa, Pereira, Liu, Koster, Kim, Zhang, Ebrahimi, Talukdar, Zheng, Poklukar, Mikhalap, Johnson, Vijayakumar, Omernick, Dibb, Dubey, Hu, Suman, Aggarwal, Kornakov, Xia, Lowe,
  Kolganov, Xiao, Nikolaev, Hemingray, Li, Iljazi, Rybiński, Sandhu, Lu, Luong, Jenatton, Govindaraj, Hui, Li, Dulac-Arnold, Park, Wang, Modi, Pouget-Abadie, Greller, Gupta, Berry, Ramachandran, Xie, McCafferty, Wang, Gupta, Lim, Bratanič, Brock, Akolzin, Sproch, Karliner, Kim, Goedeckemeyer, Shazeer, Schmid, Calandriello, Bhatia, Choromanski, Montgomery, Dua, Ramalho, King, Gao, Nguyen, Lindner, Pitta, Johnson, Salama, Ardila, Han, Farnese, Odoom, Wang, Ding, Rink, Smith, Lehri, Cohen, Vats, He, Gopavarapu, Paszke, Patel, Gansbeke, Loher, Castro, Voitovich, von Glehn, George, Niklaus, Eaton-Rosen, Rakićević, Jue, Perel, Zhang, Bahat, Pouget, Xing, Huot, Shenoy, Bos, Coriou, Richter, Noy, Wang, Ontanon, Qin, Makarchuk, Hassabis, Li, Sharma, Venkatesan, Kemaev, Daniel, Huang, Shah, Ponce, Warren, Chen, Faruqui, Wu, Andačić, Payrits, McDuff, Hume, Cao, Tessler, Wang, Wang, Rendulic, Agustsson, Johnson, Lando, Howard, Padmanabhan, Daswani, Banino, Kilgore, Heek, Ji, Caceres, Li, Kassner, Vlaskin, Liu,
  Grills, Hou, Sukkerd, Cheon, Shetty, Markeeva, Stanczyk, Iyer, Gong, Gao, Gopalakrishnan, Blyth, Reynolds, Bhoopchand, Bilenko, Gharibian, Zayats, Faust, Singh, Ma, Jiao, Vijayanarasimhan, Aroyo, Yadav, Chakera, Kakarla, Meshram, Gregor, Botea, Senter, Jia, Kovacs, Sharma, Baur, Kang, He, Zhuo, Kostelac, Laish, Peng, O'Bryan, Kasenberg, Rao, Leurent, Zhang, Stevens, Salazar, Zhang, Lobov, Walker, Porter, Redshaw, Ke, Rao, Lee, Lam, Moffitt, Kim, Qiao, Koo, Dadashi, Song, Sundararajan, Xu, Kawamoto, Zhong, Barbu, Reddy, Verzetti, Li, Papamakarios, Klimczak-Plucińska, Cassin, Kavukcuoglu, Swavely, Vaucher, Zhao, Hemsley, Tschannen, Ge, Menghani, Yu, Ha, He, Wu, Song, Sterneck, Zinke, Calian, Marsden, Ruiz, Hessel, Gueta, Lee, Farris, Gupta, Li, Saleh, Misra, Xiao, Mendolicchio, Buttimore, Krayvanova, Nayakanti, Wiethoff, Pande, Mirhoseini, Lao, Liu, Hua, Chen, Malkov, Kalashnikov, Gupta, Audhkhasi, Zhai, Kopalle, Jain, Ofek, Meyer, Baatarsukh, Strejček, Qian, Freedman, Figueira, Sokolik, Bachem, Lin,
  Kharrat, Hidey, Xu, Duan, Li, Ersoy, Everett, Cen, Santamaria-Fernandez, Taubenfeld, Mackinnon, Deng, Zablotskaia, Viswanadha, Goel, Yates, Deng, Choy, Chen, Sinha, Mossin, Wang, Szlam, Hao, Rubenstein, Toksoz-Exley, Aperghis, Zhong, Ahn, Isard, Lacombe, Luisier, Anastasiou, Kalley, Prabhu, Dunleavy, Bijwadia, Mao-Jones, Chen, Pasumarthi, Wood, Dostmohamed, Hurley, Simsa, Parrish, Pajarskas, Harvey, Skopek, Kochinski, Rey, Rieser, Zhou, Lee, Acharya, Li, Jiang, Zhang, Gipson, Mahintorabi, Gelmi, Khajehnouri, Yeh, Lee, Matthey, Baker, Pham, Fu, Pak, Gupta, Vasconcelos, Sadovsky, Walker, Hsiao, Zochbauer, Marzoca, Velan, Zeng, Baechler, Driess, Jain, Huang, Tao, Maggs, Levine, Schneider, Gemzer, Petit, Han, Fisher, Zelle, Biles, Ie, Fadeeva, Liu, Franco, Collister, Zhang, Wang, Zhao, Kieliger, Shuster, Zhu, Gong, Chan, Sun, Basu, Zimmermann, Hayes, Bapna, Snoek, Yang, Datta, Abdallah, Kilgour, Li, Mah, Jun, Rivière, Karmarkar, Spalink, Huang, Gonzalez, Tran, Nowak, Palowitch, Chadwick, Talius, Mehta, Sellam,
  Fränken, Nicosia, He, Kini, Amos, Basu, Jobe, Shaw, Xu, Evans, Ikeda, Yan, Jin, Wang, Yadav, Labzovsky, Sampath, Ma, Schumann, Siddhant, Shah, Youssef, Agarwal, Dabney, Tonioni, Ambar, Li, Guyon, Li, Soergel, Fang, Karadzhov, Udrescu, Trinh, Raunak, Noury, Guo, Gupta, Finkelstein, Petek, Liang, Billock, Sun, Wood, Song, Yu, Matejovicova, Cohen, Andra, D'Ambrosio, Deng, Nallatamby, Songhori, Dangovski, Lampinen, Botadra, Hillier, Cao, Baddi, Kuncoro, Yoshino, Bhagatwala, Ranzato, Schaeffer, Liu, Ye, Sarvana, Nham, Kuang, Gao, Baek, Mittal, Wahid, Gergely, Ni, Feldman, Muir, Lamblin, Macherey, Dyer, Kilpatrick, Campos, Bhutani, Fort, Ahmad, Severyn, Chatziprimou, Ferludin, Dimarco, Kusupati, Heyward, Bahir, Villela, Millican, Marcus, Bahargam, Unlu, Roth, Wei, Gopal, Ghoshal, Lee, Lin, Lees, Lee, Hosseini, Fan, Neel, Wu, Altun, Cai, Piqueras, Woodward, Bissacco, Haykal, Bordbar, Sundaram, Hodkinson, Toyama, Polovets, Myers, Sinha, Levinboim, Krishnakumar, Chhaparia, Sholokhova, Gundavarapu, Jawahar, Qureshi,
  Hu, Momchev, Rahtz, Wu, S, Dhamdhere, Guo, Gupta, Eslami, Schain, Blokzijl, Welling, Orr, Bolelli, Perez-Nieves, Sirotenko, Prasad, Kar, Pigem, Terzi, Weisz, Ghosh, Mavalankar, Madeka, Daugaard, Adam, Shah, Berman, Tran, Baker, Andrejczuk, Chole, Raboshchuk, Mirzazadeh, Kagohara, Wu, Schallhart, Orlando, Wang, Rrustemi, Xiong, Liu, Vezer, Ramsden, yiin Chang, Mudgal, Li, Vieillard, Hoshen, Ahmad, Slone, Hua, Potikha, Rossini, Stritar, Prakash, Wang, Dong, Nazari, Nehoran, Tekelioglu, Li, Badola, Funkhouser, Li, Yerram, Ganeshan, Formoso, Langner, Shi, Li, Yamamori, Panda, Saade, Scarpati, Breaux, Carey, Zhou, Hsieh, Bridgers, Butryna, Gupta, Tulsyan, Woo, Eltyshev, Grathwohl, Parks, Benjamin, Panigrahy, Dodhia, Freitas, Sauer, Song, Alet, Tolins, Paduraru, Zhou, Albert, Zhang, Shu, Bansal, Nguyen, Globerson, Xiao, Manyika, Hennigan, Rong, Matak, Bakalov, Sharma, Sinopalnikov, Pierson, Roller, Brown, Gao, Fukuzawa, Ghafouri, Vassigh, Barr, Wang, Korsun, Jayaram, Ren, Zaman, Khan, Lunts, Deutsch, Uthus, Katz,
  Samsikova, Khalifa, Sethi, Sun, Tang, Alon, Luo, Yu, Nayyar, Petrini, Truong, Hellendoorn, Chinaev, Alberti, Wang, Hu, Mirrokni, Balashankar, Aharon, Mehta, Iscen, Kready, Manning, Mohananey, Chen, Tripathi, Wu, Petrovski, Hwang, Baeuml, Chandrakaladharan, Liu, Coaguila, Chen, Ma, Tafti, Tatineni, Spitz, Ye, Vicol, Rosca, Puigdomènech, Yahav, Ghemawat, Lin, Kirk, Nabulsi, Brin, Bohnet, Caluwaerts, Veerubhotla, Zheng, Dai, Petrov, Xu, Mehran, Xu, Zintgraf, Choi, Hombaiah, Thoppilan, Reddi, Lew, Li, Webster, Sawhney, Lamprou, Shakeri, Lunayach, Chen, Bagri, Salcianu, Chen, Donchev, Magister, Nørly, Rodrigues, Izo, Noga, Zou, Köppe, Zhou, Lee, Long, Eisenbud, Chen, Schenck, To, Zhong, Taropa, Truong, Levy, Martins, Zhang, Semturs, Zhang, Yakubovich, Moreno, McConnaughey, Lu, Redmond, Weerts, Bitton, Refice, Lacasse, Conmy, Tallec, Odell, Forbes-Pollard, Socala, Hoech, Kohli, Walton, Wang, Sazanovich, Zhu, Kapishnikov, Galt, Denton, Murdoch, Sikora, Mohamed, Wei, First, McConnell, Cobo, Qin, Avrahami, Balle,
  Watanabe, Louis, Kraft, Ariafar, Gu, Rives, Yoon, Rusu, Cobon-Kerr, Hahn, Luo, Yuvein, Zhu, Ahuja, Benenson, Kaufman, Yu, Hightower, Zhang, Ni, Hendricks, Wang, Yona, Jain, Barrio, Bhupatiraju, Velusamy, Dafoe, Riedel, Thomas, Yuan, Bellaiche, Panthaplackel, Kloboves, Jauhari, Akbulut, Davchev, Gladchenko, Madras, Chuklin, Hill, Yuan, Madhavan, Leonhard, Scandinaro, Chen, Niu, Douillard, Damoc, Onoe, Pedregosa, Bertsch, Leichner, Pagadora, Malmaud, Ponda, Twigg, Duzhyi, Shen, Wang, Garg, Chen, Evci, Lee, Liu, Kojima, Yamaguchi, Rajendran, Piergiovanni, Rajendran, Fornoni, Ibagon, Ragan, Khan, Blitzer, Bunner, Sun, Kosakai, Lundberg, Elue, Guu, Park, Park, Narayanaswamy, Wu, Mudigonda, Cohn, Mu, Kumar, Graesser, Zhang, Killam, Zhuang, Giménez, Jishi, Ley-Wild, Zhai, Osawa, Cedillo, Liu, Upadhyay, Sieniek, Sharma, Paine, Angelova, Addepalli, Parada, Majumder, Lamp, Kumar, Deng, Myaskovsky, Sabolić, Dudek, York, de~Chaumont~Quitry, Nie, Cattle, Gunjan, Piot, Khawaja, Bang, Wang, Khodadadeh, R, Rawlani,
  Powell, Lee, Griesser, Oh, Magalhaes, Li, Tokumine, Vogel, Hsu, BC, Jindal, Cohen, Yang, Yuan, de~Cesare, Bruguier, Xu, Roy, Jacovi, Belov, Arya, Meadowlark, Cohen-Ganor, Ye, Morris-Suzuki, Banzal, Song, Ponnuramu, Zhang, Scrivener, Zaiem, Rochman, Han, Ghazi, Lee, Drath, Suo, Girgis, Shenoy, Nguyen, Eck, Gupta, Yan, Carreira, Gulati, Sang, Mirylenka, Cooney, Chou, Ling, Fan, Coleman, Tubone, Kumar, Baldridge, Hernandez-Campos, Lazaridou, Besley, Yona, Bulut, Wellens, Pierigiovanni, George, Green, Han, Tao, Clark, You, Abdolmaleki, Fu, Chen, Chaugule, Chandorkar, Rahman, Thompson, Koanantakool, Bernico, Ren, Vlasov, Vassilvitskii, Kula, Liang, Kim, Huang, Ye, Lepikhin, and Helmholz}]{comanici2025gemini}
Gheorghe Comanici, Eric Bieber, Mike Schaekermann, Ice Pasupat, Noveen Sachdeva, Inderjit Dhillon, Marcel Blistein, Ori Ram, Dan Zhang, Evan Rosen, Luke Marris, Sam Petulla, Colin Gaffney, Asaf Aharoni, Nathan Lintz, Tiago~Cardal Pais, Henrik Jacobsson, Idan Szpektor, Nan-Jiang Jiang, and 3416 others. 2025.
\newblock \href {https://arxiv.org/abs/2507.06261} {Gemini 2.5: Pushing the frontier with advanced reasoning, multimodality, long context, and next generation agentic capabilities}.
\newblock \emph{Preprint}, arXiv:2507.06261.

\bibitem[{Deng et~al.(2023)Deng, Gu, Zheng, Chen, Stevens, Wang, Sun, and Su}]{deng2023mind2web}
Xiang Deng, Yu~Gu, Boyuan Zheng, Shijie Chen, Samuel Stevens, Boshi Wang, Huan Sun, and Yu~Su. 2023.
\newblock \href {https://arxiv.org/abs/2306.06070} {Mind2web: Towards a generalist agent for the web}.
\newblock \emph{Preprint}, arXiv:2306.06070.

\bibitem[{Feng et~al.(2025)Feng, Xue, Liu, and An}]{feng2025groupingrouppolicyoptimizationllm}
Lang Feng, Zhenghai Xue, Tingcong Liu, and Bo~An. 2025.
\newblock \href {https://arxiv.org/abs/2505.10978} {Group-in-group policy optimization for llm agent training}.
\newblock \emph{Preprint}, arXiv:2505.10978.

\bibitem[{Feng et~al.(2024)Feng, He, Huang, Lin, Zhang, Zhang, and Li}]{feng2024agile}
Peiyuan Feng, Yichen He, Guanhua Huang, Yuan Lin, Hanchong Zhang, Yuchen Zhang, and Hang Li. 2024.
\newblock \href {https://arxiv.org/abs/2405.14751} {Agile: A novel reinforcement learning framework of llm agents}.
\newblock \emph{Preprint}, arXiv:2405.14751.

\bibitem[{Florensa et~al.(2018)Florensa, Held, Geng, and Abbeel}]{florensa2018automatic}
Carlos Florensa, David Held, Xinyang Geng, and Pieter Abbeel. 2018.
\newblock \href {https://arxiv.org/abs/1705.06366} {Automatic goal generation for reinforcement learning agents}.
\newblock \emph{Preprint}, arXiv:1705.06366.

\bibitem[{Gu et~al.(2025)Gu, Zhang, Ning, Zheng, Gou, Xue, Chang, Srivastava, Xie, Qi, Sun, and Su}]{gu2024webdreamer}
Yu~Gu, Kai Zhang, Yuting Ning, Boyuan Zheng, Boyu Gou, Tianci Xue, Cheng Chang, Sanjari Srivastava, Yanan Xie, Peng Qi, Huan Sun, and Yu~Su. 2025.
\newblock \href {https://arxiv.org/abs/2411.06559} {Is your llm secretly a world model of the internet? model-based planning for web agents}.
\newblock \emph{Preprint}, arXiv:2411.06559.

\bibitem[{Guo et~al.(2025)Guo, Deng, Cheng, and Tang}]{guo2025g2rpoa}
Yongxin Guo, Wenbo Deng, Zhenglin Cheng, and Xiaoying Tang. 2025.
\newblock \href {https://arxiv.org/abs/2508.13023} {G$^2$rpo-a: Guided group relative policy optimization with adaptive guidance}.
\newblock \emph{Preprint}, arXiv:2508.13023.

\bibitem[{Huang et~al.(2022)Huang, Abbeel, Pathak, and Mordatch}]{huang2022language}
Wenlong Huang, Pieter Abbeel, Deepak Pathak, and Igor Mordatch. 2022.
\newblock \href {https://arxiv.org/abs/2201.07207} {Language models as zero-shot planners: Extracting actionable knowledge for embodied agents}.
\newblock \emph{Preprint}, arXiv:2201.07207.

\bibitem[{Huang et~al.(2026)Huang, Cheng, Qiu, Wang, Xu, Ponti, and Titov}]{huang2025prefixrft}
Zeyu Huang, Tianhao Cheng, Zihan Qiu, Zili Wang, Yinghui Xu, Edoardo~M. Ponti, and Ivan Titov. 2026.
\newblock \href {https://arxiv.org/abs/2507.01679} {Blending supervised and reinforcement fine-tuning with prefix sampling}.
\newblock \emph{Preprint}, arXiv:2507.01679.

\bibitem[{Lai et~al.(2024)Lai, Liu, Iong, Yao, Chen, Shen, Yu, Zhang, Zhang, Dong, and Tang}]{lai2024autowebglm}
Hanyu Lai, Xiao Liu, Iat~Long Iong, Shuntian Yao, Yuxuan Chen, Pengbo Shen, Hao Yu, Hanchen Zhang, Xiaohan Zhang, Yuxiao Dong, and Jie Tang. 2024.
\newblock \href {https://arxiv.org/abs/2404.03648} {Autowebglm: A large language model-based web navigating agent}.
\newblock \emph{Preprint}, arXiv:2404.03648.

\bibitem[{Li et~al.(2026)Li, Zhao, Cheng, Lin, Peng, Li, Wang, Dai, Li, Wang, Shi, Zhao, Sun, Ge, Zhang, Lu, Xiao, Zhuang, and Shen}]{li2026spatialevoselfevolvingspatialintelligence}
Dinging Li, Yingxiu Zhao, Xinrui Cheng, Kangheng Lin, Hongbo Peng, Hongxing Li, Zixuan Wang, Yuhong Dai, Haodong Li, Jia Wang, Yukang Shi, Liang Zhao, Jianjian Sun, Zheng Ge, Xiangyu Zhang, Weiming Lu, Jun Xiao, Yueting Zhuang, and Yongliang Shen. 2026.
\newblock \href {https://arxiv.org/abs/2604.14144} {Spatialevo: Self-evolving spatial intelligence via deterministic geometric environments}.
\newblock \emph{Preprint}, arXiv:2604.14144.

\bibitem[{Li et~al.(2025)Li, Sun, Zhao, Min, Zeng, Wu, Cai, Wang, Yin, Chen, and Deng}]{li2025seele}
Ziheng Li, Zexu Sun, Jinman Zhao, Erxue Min, Yongcheng Zeng, Hui Wu, Hengyi Cai, Shuaiqiang Wang, Dawei Yin, Xu~Chen, and Zhi-Hong Deng. 2025.
\newblock \href {https://arxiv.org/abs/2509.06923} {Staying in the sweet spot: Responsive reasoning evolution via capability-adaptive hint scaffolding}.
\newblock \emph{Preprint}, arXiv:2509.06923.

\bibitem[{Lian et~al.(2026{\natexlab{a}})Lian, Chen, Yu, Duan, Liu, Liu, Fu, Luan, Qu, Xia, and Wang}]{lian2026uimopdmultiplatformonpolicydistillation}
Niu Lian, Tongbo Chen, Zhehao Yu, Chengzhen Duan, Fazhan Liu, Hui Liu, Pei Fu, Jian Luan, Heng Qu, Shu-Tao Xia, and Jinpeng Wang. 2026{\natexlab{a}}.
\newblock \href {https://arxiv.org/abs/2607.04425} {Ui-mopd: Multi-platform on-policy distillation for unified gui agents}.
\newblock \emph{Preprint}, arXiv:2607.04425.

\bibitem[{Lian et~al.(2026{\natexlab{b}})Lian, Wang, Yao, Wang, Chen, Wang, Zhang, and Xia}]{lian2026verbatim}
Niu Lian, Yuting Wang, Hanshu Yao, Jinpeng Wang, Bin Chen, Yaowei Wang, Min Zhang, and Shu-Tao Xia. 2026{\natexlab{b}}.
\newblock From verbatim to gist: Distilling pyramidal multimodal memory via semantic information bottleneck for long-horizon video agents.
\newblock In \emph{Proceedings of the 64th Annual Meeting of the Association for Computational Linguistics (Volume 1: Long Papers)}, pages 11601--11617.

\bibitem[{Lu et~al.(2026{\natexlab{a}})Lu, Yao, Han, Wang, Wu, Gu, Cai, Lu, Xiao, Zhuang, and Shen}]{lu2026sdar}
Zhengxi Lu, Zhiyuan Yao, Zhuowen Han, Zi-Han Wang, Jinyang Wu, Qi~Gu, Xunliang Cai, Weiming Lu, Jun Xiao, Yueting Zhuang, and Yongliang Shen. 2026{\natexlab{a}}.
\newblock \href {https://arxiv.org/abs/2605.15155} {Self-distilled agentic reinforcement learning}.
\newblock \emph{Preprint}, arXiv:2605.15155.

\bibitem[{Lu et~al.(2026{\natexlab{b}})Lu, Yao, Wu, Han, Gu, Cai, Lu, Xiao, Zhuang, and Shen}]{lu2026skill0}
Zhengxi Lu, Zhiyuan Yao, Jinyang Wu, Chengcheng Han, Qi~Gu, Xunliang Cai, Weiming Lu, Jun Xiao, Yueting Zhuang, and Yongliang Shen. 2026{\natexlab{b}}.
\newblock \href {https://arxiv.org/abs/2604.02268} {Skill0: In-context agentic reinforcement learning for skill internalization}.
\newblock \emph{Preprint}, arXiv:2604.02268.

\bibitem[{OpenAI et~al.(2024)OpenAI, :, Hurst, Lerer, Goucher, Perelman, Ramesh, Clark, Ostrow, Welihinda, Hayes, Radford, Mądry, Baker-Whitcomb, Beutel, Borzunov, Carney, Chow, Kirillov, Nichol, Paino, Renzin, Passos, Kirillov, Christakis, Conneau, Kamali, Jabri, Moyer, Tam, Crookes, Tootoochian, Tootoonchian, Kumar, Vallone, Karpathy, Braunstein, Cann, Codispoti, Galu, Kondrich, Tulloch, Mishchenko, Baek, Jiang, Pelisse, Woodford, Gosalia, Dhar, Pantuliano, Nayak, Oliver, Zoph, Ghorbani, Leimberger, Rossen, Sokolowsky, Wang, Zweig, Hoover, Samic, McGrew, Spero, Giertler, Cheng, Lightcap, Walkin, Quinn, Guarraci, Hsu, Kellogg, Eastman, Lugaresi, Wainwright, Bassin, Hudson, Chu, Nelson, Li, Shern, Conger, Barette, Voss, Ding, Lu, Zhang, Beaumont, Hallacy, Koch, Gibson, Kim, Choi, McLeavey, Hesse, Fischer, Winter, Czarnecki, Jarvis, Wei, Koumouzelis, Sherburn, Kappler, Levin, Levy, Carr, Farhi, Mely, Robinson, Sasaki, Jin, Valladares, Tsipras, Li, Nguyen, Findlay, Oiwoh, Wong, Asdar, Proehl, Yang, Antonow,
  Kramer, Peterson, Sigler, Wallace, Brevdo, Mays, Khorasani, Such, Raso, Zhang, von Lohmann, Sulit, Goh, Oden, Salmon, Starace, Brockman, Salman, Bao, Hu, Wong, Wang, Schmidt, Whitney, Jun, Kirchner, de~Oliveira~Pinto, Ren, Chang, Chung, Kivlichan, O'Connell, O'Connell, Osband, Silber, Sohl, Okuyucu, Lan, Kostrikov, Sutskever, Kanitscheider, Gulrajani, Coxon, Menick, Pachocki, Aung, Betker, Crooks, Lennon, Kiros, Leike, Park, Kwon, Phang, Teplitz, Wei, Wolfe, Chen, Harris, Varavva, Lee, Shieh, Lin, Yu, Weng, Tang, Yu, Jang, Candela, Beutler, Landers, Parish, Heidecke, Schulman, Lachman, McKay, Uesato, Ward, Kim, Huizinga, Sitkin, Kraaijeveld, Gross, Kaplan, Snyder, Achiam, Jiao, Lee, Zhuang, Harriman, Fricke, Hayashi, Singhal, Shi, Karthik, Wood, Rimbach, Hsu, Nguyen, Gu-Lemberg, Button, Liu, Howe, Muthukumar, Luther, Ahmad, Kai, Itow, Workman, Pathak, Chen, Jing, Guy, Fedus, Zhou, Mamitsuka, Weng, McCallum, Held, Ouyang, Feuvrier, Zhang, Kondraciuk, Kaiser, Hewitt, Metz, Doshi, Aflak, Simens, Boyd,
  Thompson, Dukhan, Chen, Gray, Hudnall, Zhang, Aljubeh, Litwin, Zeng, Johnson, Shetty, Gupta, Shah, Yatbaz, Yang, Zhong, Glaese, Chen, Janner, Lampe, Petrov, Wu, Wang, Fradin, Pokrass, Castro, de~Castro, Pavlov, Brundage, Wang, Khan, Murati, Bavarian, Lin, Yesildal, Soto, Gimelshein, Cone, Staudacher, Summers, LaFontaine, Chowdhury, Ryder, Stathas, Turley, Tezak, Felix, Kudige, Keskar, Deutsch, Bundick, Puckett, Nachum, Okelola, Boiko, Murk, Jaffe, Watkins, Godement, Campbell-Moore, Chao, McMillan, Belov, Su, Bak, Bakkum, Deng, Dolan, Hoeschele, Welinder, Tillet, Pronin, Tillet, Dhariwal, Yuan, Dias, Lim, Arora, Troll, Lin, Lopes, Puri, Miyara, Leike, Gaubert, Zamani, Wang, Donnelly, Honsby, Smith, Sahai, Ramchandani, Huet, Carmichael, Zellers, Chen, Chen, Nigmatullin, Cheu, Jain, Altman, Schoenholz, Toizer, Miserendino, Agarwal, Culver, Ethersmith, Gray, Grove, Metzger, Hermani, Jain, Zhao, Wu, Jomoto, Wu, Shuaiqi, Xia, Phene, Papay, Narayanan, Coffey, Lee, Hall, Balaji, Broda, Stramer, Xu, Gogineni,
  Christianson, Sanders, Patwardhan, Cunninghman, Degry, Dimson, Raoux, Shadwell, Zheng, Underwood, Markov, Sherbakov, Rubin, Stasi, Kaftan, Heywood, Peterson, Walters, Eloundou, Qi, Moeller, Monaco, Kuo, Fomenko, Chang, Zheng, Zhou, Manassra, Sheu, Zaremba, Patil, Qian, Kim, Cheng, Zhang, He, Zhang, Jin, Dai, and Malkov}]{hurst2024gpt}
OpenAI, :, Aaron Hurst, Adam Lerer, Adam~P. Goucher, Adam Perelman, Aditya Ramesh, Aidan Clark, AJ~Ostrow, Akila Welihinda, Alan Hayes, Alec Radford, Aleksander Mądry, Alex Baker-Whitcomb, Alex Beutel, Alex Borzunov, Alex Carney, Alex Chow, Alex Kirillov, and 401 others. 2024.
\newblock \href {https://arxiv.org/abs/2410.21276} {Gpt-4o system card}.
\newblock \emph{Preprint}, arXiv:2410.21276.

\bibitem[{Pan et~al.(2026)Pan, Yan, Wang, Zhang, Hou, Zhang, Lu, Xiao, and Shen}]{pan-etal-2026-coverrl}
Teng Pan, Yuchen Yan, Zixuan Wang, Ruiqing Zhang, Guiyang Hou, Wenqi Zhang, Weiming Lu, Jun Xiao, and Yongliang Shen. 2026.
\newblock \href {https://doi.org/10.18653/v1/2026.acl-long.1376} {{C}o{V}er{RL}: Breaking the consensus trap in label-free reasoning via generator-verifier co-evolution}.
\newblock In \emph{Proceedings of the 64th Annual Meeting of the {A}ssociation for {C}omputational {L}inguistics (Volume 1: Long Papers)}, pages 29833--29853, San Diego, California, United States. Association for Computational Linguistics.

\bibitem[{Putta et~al.(2024)Putta, Mills, Garg, Motwani, Finn, Garg, and Rafailov}]{putta2024agentq}
Pranav Putta, Edmund Mills, Naman Garg, Sumeet Motwani, Chelsea Finn, Divyansh Garg, and Rafael Rafailov. 2024.
\newblock \href {https://arxiv.org/abs/2408.07199} {Agent q: Advanced reasoning and learning for autonomous ai agents}.
\newblock \emph{Preprint}, arXiv:2408.07199.

\bibitem[{Qi et~al.(2025)Qi, Liu, Iong, Lai, Sun, Zhao, Yang, Yang, Sun, Yao, Zhang, Xu, Tang, and Dong}]{qi2025webrl}
Zehan Qi, Xiao Liu, Iat~Long Iong, Hanyu Lai, Xueqiao Sun, Wenyi Zhao, Yu~Yang, Xinyue Yang, Jiadai Sun, Shuntian Yao, Tianjie Zhang, Wei Xu, Jie Tang, and Yuxiao Dong. 2025.
\newblock \href {https://arxiv.org/abs/2411.02337} {Webrl: Training llm web agents via self-evolving online curriculum reinforcement learning}.
\newblock \emph{Preprint}, arXiv:2411.02337.

\bibitem[{Qwen et~al.(2025)Qwen, :, Yang, Yang, Zhang, Hui, Zheng, Yu, Li, Liu, Huang, Wei, Lin, Yang, Tu, Zhang, Yang, Yang, Zhou, Lin, Dang, Lu, Bao, Yang, Yu, Li, Xue, Zhang, Zhu, Men, Lin, Li, Tang, Xia, Ren, Ren, Fan, Su, Zhang, Wan, Liu, Cui, Zhang, and Qiu}]{yang2024qwen25}
Qwen, :, An~Yang, Baosong Yang, Beichen Zhang, Binyuan Hui, Bo~Zheng, Bowen Yu, Chengyuan Li, Dayiheng Liu, Fei Huang, Haoran Wei, Huan Lin, Jian Yang, Jianhong Tu, Jianwei Zhang, Jianxin Yang, Jiaxi Yang, Jingren Zhou, and 25 others. 2025.
\newblock \href {https://arxiv.org/abs/2412.15115} {Qwen2.5 technical report}.
\newblock \emph{Preprint}, arXiv:2412.15115.

\bibitem[{Shao et~al.(2024)Shao, Wang, Zhu, Xu, Song, Bi, Zhang, Zhang, Li, Wu, and Guo}]{shao2024deepseekmathpushinglimitsmathematical}
Zhihong Shao, Peiyi Wang, Qihao Zhu, Runxin Xu, Junxiao Song, Xiao Bi, Haowei Zhang, Mingchuan Zhang, Y.~K. Li, Y.~Wu, and Daya Guo. 2024.
\newblock \href {https://arxiv.org/abs/2402.03300} {Deepseekmath: Pushing the limits of mathematical reasoning in open language models}.
\newblock \emph{Preprint}, arXiv:2402.03300.

\bibitem[{Shen et~al.(2023)Shen, Song, Tan, Li, Lu, and Zhuang}]{NEURIPS2023_77c33e6a}
Yongliang Shen, Kaitao Song, Xu~Tan, Dongsheng Li, Weiming Lu, and Yueting Zhuang. 2023.
\newblock \href {https://doi.org/10.52202/075280-1657} {Hugginggpt: Solving ai tasks with chatgpt and its friends in hugging face}.
\newblock In \emph{Advances in Neural Information Processing Systems}, volume~36, pages 38154--38180. Curran Associates, Inc.

\bibitem[{Sheng et~al.(2025)Sheng, Zhang, Ye, Wu, Zhang, Zhang, Peng, Lin, and Wu}]{sheng2024hybridflow}
Guangming Sheng, Chi Zhang, Zilingfeng Ye, Xibin Wu, Wang Zhang, Ru~Zhang, Yanghua Peng, Haibin Lin, and Chuan Wu. 2025.
\newblock \href {https://doi.org/10.1145/3689031.3696075} {Hybridflow: A flexible and efficient rlhf framework}.
\newblock In \emph{Proceedings of the Twentieth European Conference on Computer Systems}, EuroSys '25, page 1279–1297, New York, NY, USA. Association for Computing Machinery.

\bibitem[{Shinn et~al.(2023)Shinn, Cassano, Berman, Gopinath, Narasimhan, and Yao}]{shinn2023reflexionlanguageagentsverbal}
Noah Shinn, Federico Cassano, Edward Berman, Ashwin Gopinath, Karthik Narasimhan, and Shunyu Yao. 2023.
\newblock \href {https://arxiv.org/abs/2303.11366} {Reflexion: Language agents with verbal reinforcement learning}.
\newblock \emph{Preprint}, arXiv:2303.11366.

\bibitem[{Shridhar et~al.(2021)Shridhar, Yuan, Côté, Bisk, Trischler, and Hausknecht}]{ALFWorld20}
Mohit Shridhar, Xingdi Yuan, Marc-Alexandre Côté, Yonatan Bisk, Adam Trischler, and Matthew Hausknecht. 2021.
\newblock \href {https://arxiv.org/abs/2010.03768} {Alfworld: Aligning text and embodied environments for interactive learning}.
\newblock \emph{Preprint}, arXiv:2010.03768.

\bibitem[{Song et~al.(2024)Song, Yin, Yue, Huang, Li, and Lin}]{song2024eto}
Yifan Song, Da~Yin, Xiang Yue, Jie Huang, Sujian Li, and Bill~Yuchen Lin. 2024.
\newblock Trial and error: Exploration-based trajectory optimization for {LLM} agents.
\newblock In \emph{Proceedings of the 62nd Annual Meeting of the Association for Computational Linguistics (ACL)}.

\bibitem[{Su et~al.(2025)Su, Guan, Gu, Huang, and Wang}]{su2025trapo}
Mingyu Su, Jian Guan, Yuxian Gu, Minlie Huang, and Hongning Wang. 2025.
\newblock \href {https://arxiv.org/abs/2512.17636} {Trust-region adaptive policy optimization}.
\newblock \emph{Preprint}, arXiv:2512.17636.

\bibitem[{Tu et~al.(2026)Tu, Li, Huang, and Xu}]{tu2026consensus}
Geng Tu, Dingming Li, Jun Huang, and Ruifeng Xu. 2026.
\newblock Consensus-driven multi-agent cognitive reasoning for enhancing the emotional intelligence of large language models.
\newblock In \emph{Proceedings of the AAAI Conference on Artificial Intelligence}, volume~40, pages 17751--17759.

\bibitem[{Wang et~al.(2025{\natexlab{a}})Wang, Leong, Wang, Wang, and Li}]{wang2025sparl}
Hanlin Wang, Chak~Tou Leong, Jiashuo Wang, Jian Wang, and Wenjie Li. 2025{\natexlab{a}}.
\newblock \href {https://arxiv.org/abs/2505.20732} {Spa-rl: Reinforcing llm agents via stepwise progress attribution}.
\newblock \emph{Preprint}, arXiv:2505.20732.

\bibitem[{Wang et~al.(2025{\natexlab{b}})Wang, Han, Jiang, Li, Liang, Jiang, Dai, Ma, Yu, and Xiao}]{wang2025hint}
Xinyi Wang, Jinyi Han, Zishang Jiang, Tingyun Li, Jiaqing Liang, Sihang Jiang, Zhaoqian Dai, Shuguang Ma, Fei Yu, and Yanghua Xiao. 2025{\natexlab{b}}.
\newblock \href {https://arxiv.org/abs/2510.09388} {Hint: Helping ineffective rollouts navigate towards effectiveness}.
\newblock \emph{Preprint}, arXiv:2510.09388.

\bibitem[{Wang et~al.(2025{\natexlab{c}})Wang, Li, Li, Chen, Yan, Zhang, Shen, Lu, Xiao, and Zhuang}]{wang2025omniear}
Zixuan Wang, Dingming Li, Hongxing Li, Shuo Chen, Yuchen Yan, Wenqi Zhang, Yongliang Shen, Weiming Lu, Jun Xiao, and Yueting Zhuang. 2025{\natexlab{c}}.
\newblock \href {https://arxiv.org/abs/2508.05614} {Omniear: Benchmarking agent reasoning in embodied tasks}.
\newblock \emph{Preprint}, arXiv:2508.05614.

\bibitem[{Wang et~al.(2026)Wang, Yan, Li, Pan, Li, Zhang, Lu, Xiao, Zhuang, and Shen}]{wang2026beacon}
Zixuan Wang, Yuchen Yan, Hongxing Li, Teng Pan, Dingming Li, Ruiqing Zhang, Weiming Lu, Jun Xiao, Yueting Zhuang, and Yongliang Shen. 2026.
\newblock \href {https://arxiv.org/abs/2605.06078} {Milestone-guided policy learning for long-horizon language agents}.
\newblock \emph{Preprint}, arXiv:2605.06078.

\bibitem[{Xi et~al.(2024{\natexlab{a}})Xi, Chen, Hong, Jin, Zheng, He, Ding, Liu, Guo, Wang, Guo, Shen, Fan, Zhou, Dou, Wang, Zhang, Sun, Gui, Zhang, and Huang}]{xi2024r3}
Zhiheng Xi, Wenxiang Chen, Boyang Hong, Senjie Jin, Rui Zheng, Wei He, Yiwen Ding, Shichun Liu, Xin Guo, Junzhe Wang, Honglin Guo, Wei Shen, Xiaoran Fan, Yuhao Zhou, Shihan Dou, Xiao Wang, Xinbo Zhang, Peng Sun, Tao Gui, and 2 others. 2024{\natexlab{a}}.
\newblock \href {https://arxiv.org/abs/2402.05808} {Training large language models for reasoning through reverse curriculum reinforcement learning}.
\newblock \emph{Preprint}, arXiv:2402.05808.

\bibitem[{Xi et~al.(2024{\natexlab{b}})Xi, Ding, Chen, Hong, Guo, Wang, Yang, Liao, Guo, He, Gao, Chen, Zheng, Zou, Gui, Zhang, Qiu, Huang, Wu, and Jiang}]{xi2025agentgym}
Zhiheng Xi, Yiwen Ding, Wenxiang Chen, Boyang Hong, Honglin Guo, Junzhe Wang, Dingwen Yang, Chenyang Liao, Xin Guo, Wei He, Songyang Gao, Lu~Chen, Rui Zheng, Yicheng Zou, Tao Gui, Qi~Zhang, Xipeng Qiu, Xuanjing Huang, Zuxuan Wu, and Yu-Gang Jiang. 2024{\natexlab{b}}.
\newblock \href {https://arxiv.org/abs/2406.04151} {Agentgym: Evolving large language model-based agents across diverse environments}.
\newblock \emph{Preprint}, arXiv:2406.04151.

\bibitem[{Xiang et~al.(2024)Xiang, Shen, Zhang, and Nguyen}]{xiang2024retrospex}
Yufei Xiang, Yiqun Shen, Yeqin Zhang, and Cam-Tu Nguyen. 2024.
\newblock {Retrospex}: Language agent meets offline reinforcement learning critic.
\newblock In \emph{Proceedings of the 2024 Conference on Empirical Methods in Natural Language Processing (EMNLP)}.

\bibitem[{Xiong et~al.(2024)Xiong, Song, Zhao, Wu, Wang, Wang, Li, Peng, and Li}]{xiong2024ipr}
Weimin Xiong, Yifan Song, Xiutian Zhao, Wenhao Wu, Xun Wang, Ke~Wang, Cheng Li, Wei Peng, and Sujian Li. 2024.
\newblock \href {https://arxiv.org/abs/2406.11176} {Watch every step! llm agent learning via iterative step-level process refinement}.
\newblock \emph{Preprint}, arXiv:2406.11176.

\bibitem[{Yao et~al.(2023{\natexlab{a}})Yao, Chen, Yang, and Narasimhan}]{Yao2022WebShopTS}
Shunyu Yao, Howard Chen, John Yang, and Karthik Narasimhan. 2023{\natexlab{a}}.
\newblock \href {https://arxiv.org/abs/2207.01206} {Webshop: Towards scalable real-world web interaction with grounded language agents}.
\newblock \emph{Preprint}, arXiv:2207.01206.

\bibitem[{Yao et~al.(2023{\natexlab{b}})Yao, Zhao, Yu, Du, Shafran, Narasimhan, and Cao}]{yao2023reactsynergizingreasoningacting}
Shunyu Yao, Jeffrey Zhao, Dian Yu, Nan Du, Izhak Shafran, Karthik Narasimhan, and Yuan Cao. 2023{\natexlab{b}}.
\newblock \href {https://arxiv.org/abs/2210.03629} {React: Synergizing reasoning and acting in language models}.
\newblock \emph{Preprint}, arXiv:2210.03629.

\bibitem[{Yuan et~al.(2025)Yuan, Chen, Xi, Ye, Du, and Chen}]{yuan2025agentr}
Siyu Yuan, Zehui Chen, Zhiheng Xi, Junjie Ye, Zhengyin Du, and Jiecao Chen. 2025.
\newblock \href {https://arxiv.org/abs/2501.11425} {Agent-r: Training language model agents to reflect via iterative self-training}.
\newblock \emph{Preprint}, arXiv:2501.11425.

\bibitem[{Zeng et~al.(2023)Zeng, Liu, Lu, Wang, Liu, Dong, and Tang}]{zeng2024agenttuning}
Aohan Zeng, Mingdao Liu, Rui Lu, Bowen Wang, Xiao Liu, Yuxiao Dong, and Jie Tang. 2023.
\newblock \href {https://arxiv.org/abs/2310.12823} {Agenttuning: Enabling generalized agent abilities for llms}.
\newblock \emph{Preprint}, arXiv:2310.12823.

\bibitem[{Zhang et~al.(2025{\natexlab{a}})Zhang, Lv, Li, Wang, Wang, Hu, and Yan}]{zhang2025stephint}
Kaiyi Zhang, Ang Lv, Jinpeng Li, Yongbo Wang, Feng Wang, Haoyuan Hu, and Rui Yan. 2025{\natexlab{a}}.
\newblock \href {https://arxiv.org/abs/2507.02841} {Stephint: Multi-level stepwise hints enhance reinforcement learning to reason}.
\newblock \emph{Preprint}, arXiv:2507.02841.

\bibitem[{Zhang et~al.(2026{\natexlab{a}})Zhang, Yao, Qin, Xu, Zhu, Yu, Wang, Tang, Gu, Deng, and Chen}]{Zhang_2026}
Ningyu Zhang, Yunzhi Yao, Jiaxin Qin, Haoming Xu, Yuqi Zhu, Zeping Yu, Mengru Wang, Yuqi Tang, Jia-Chen Gu, Shumin Deng, and Huajun Chen. 2026{\natexlab{a}}.
\newblock \href {https://doi.org/10.1038/s42256-026-01276-y} {Towards principled knowledge editing methods for large language model reasoning}.
\newblock \emph{Nature Machine Intelligence}, 8(8):1189–1200.

\bibitem[{Zhang et~al.(2026{\natexlab{b}})Zhang, Wu, Zhu, Tan, Yu, He, and Jia}]{zhang2025scafgrpo}
Xichen Zhang, Sitong Wu, Yinghao Zhu, Haoru Tan, Shaozuo Yu, Ziyi He, and Jiaya Jia. 2026{\natexlab{b}}.
\newblock \href {https://arxiv.org/abs/2510.19807} {Scaf-grpo: Scaffolded group relative policy optimization for enhancing llm reasoning}.
\newblock \emph{Preprint}, arXiv:2510.19807.

\bibitem[{Zhang et~al.(2025{\natexlab{b}})Zhang, Huang, Li, Ni, Chen, and Oymak}]{zhang2025bread}
Xuechen Zhang, Zijian Huang, Yingcong Li, Chenshun Ni, Jiasi Chen, and Samet Oymak. 2025{\natexlab{b}}.
\newblock \href {https://arxiv.org/abs/2506.17211} {Bread: Branched rollouts from expert anchors bridge sft \& rl for reasoning}.
\newblock \emph{Preprint}, arXiv:2506.17211.

\bibitem[{Zhang et~al.(2025{\natexlab{c}})Zhang, Zheng, Wu, Zhang, Lin, Yu, Liu, Zhou, and Lin}]{zhang2025prmlessons}
Zhenru Zhang, Chujie Zheng, Yangzhen Wu, Beichen Zhang, Runji Lin, Bowen Yu, Dayiheng Liu, Jingren Zhou, and Junyang Lin. 2025{\natexlab{c}}.
\newblock \href {https://arxiv.org/abs/2501.07301} {The lessons of developing process reward models in mathematical reasoning}.
\newblock \emph{Preprint}, arXiv:2501.07301.

\bibitem[{Zhang et~al.(2025{\natexlab{d}})Zhang, Chen, Li, Tu, and Li}]{zhang2025rlvmrreinforcementlearningverifiable}
Zijing Zhang, Ziyang Chen, Mingxiao Li, Zhaopeng Tu, and Xiaolong Li. 2025{\natexlab{d}}.
\newblock \href {https://arxiv.org/abs/2507.22844} {Rlvmr: Reinforcement learning with verifiable meta-reasoning rewards for robust long-horizon agents}.
\newblock \emph{Preprint}, arXiv:2507.22844.

\bibitem[{Zheng et~al.(2026)Zheng, Mo, Ma, Lin, Zhao, Zhu, Lou, Wang, Wang, Liu, Zhang, Yu, and Zhang}]{zheng2026admire}
Congmin Zheng, Xiaoyun Mo, Xinbei Ma, Qiqiang Lin, Yin Zhao, Jiachen Zhu, Xingyu Lou, Jun Wang, Zhaoxiang Wang, Weiwen Liu, Zhuosheng Zhang, Yong Yu, and Weinan Zhang. 2026.
\newblock \href {https://arxiv.org/abs/2602.11524} {Adaptive milestone reward for gui agents}.
\newblock \emph{Preprint}, arXiv:2602.11524.

\bibitem[{Zhou et~al.(2024{\natexlab{a}})Zhou, Xu, Zhu, Zhou, Lo, Sridhar, Cheng, Ou, Bisk, Fried, Alon, and Neubig}]{zhou2023webarena}
Shuyan Zhou, Frank~F. Xu, Hao Zhu, Xuhui Zhou, Robert Lo, Abishek Sridhar, Xianyi Cheng, Tianyue Ou, Yonatan Bisk, Daniel Fried, Uri Alon, and Graham Neubig. 2024{\natexlab{a}}.
\newblock \href {https://arxiv.org/abs/2307.13854} {Webarena: A realistic web environment for building autonomous agents}.
\newblock \emph{Preprint}, arXiv:2307.13854.

\bibitem[{Zhou et~al.(2024{\natexlab{b}})Zhou, Zanette, Pan, Levine, and Kumar}]{zhou2024archer}
Yifei Zhou, Andrea Zanette, Jiayi Pan, Sergey Levine, and Aviral Kumar. 2024{\natexlab{b}}.
\newblock \href {https://arxiv.org/abs/2402.19446} {Archer: Training language model agents via hierarchical multi-turn rl}.
\newblock \emph{Preprint}, arXiv:2402.19446.

\end{thebibliography}
\end{document}